\documentclass[10pt,twocolumn,letterpaper]{article}

\usepackage{siunitx}

\usepackage{tikz}
\usepackage{pgfplots}
\usepgfplotslibrary{fillbetween}
\usepackage{iccv}
\usepackage{times}
\usepackage{epsfig}
\usepackage{graphicx}
\usepackage{amsmath}
\usepackage{amssymb}
\usepackage{algorithm}
\usepackage[noend]{algpseudocode}

\usepackage[pagebackref=true,breaklinks=true,letterpaper=true,colorlinks,bookmarks=false]{hyperref}

  \iccvfinalcopy
\begin{document}

%%%%%%%%% TITLE
\title{Learning Variations in Human Motion via Mix-and-Match Perturbation}

\author{
Sadegh Aliakbarian \thanks{Work partially done during internship with Qualcomm AI Research.}  \thanks{Equal contribution.}\\ANU, ACRV, CSIRO\\
{\tt\scriptsize sadegh.aliakbarian@anu.edu.au}
\and
Fatemeh Sadat Saleh \footnotemark[2]\\ANU, ACRV\\
{\tt\scriptsize fatemehsadat.saleh@anu.edu.au}
\and
Mathieu Salzmann\\CVLab-EPFL\\
{\tt\scriptsize mathieu.salzmann@epfl.ch}
\and
Lars Petersson\\CSIRO, ANU\\
{\tt\scriptsize lars.petersson@data61.csiro.au}
\and
Stephen Gould\\ANU, ACRV\\
{\tt\scriptsize stephen.gould@anu.edu.au}
\and
Amirhossein Habibian\\Qualcomm AI Research\\
{\tt\scriptsize ahabibia@qti.qualcomm.com}
}

\maketitle

\begin{abstract}
Human motion prediction is a stochastic process: Given an observed sequence of poses, multiple future motions are plausible. Existing approaches to modeling this stochasticity typically combine a random noise vector with information about the previous poses. This combination, however, is done in a deterministic manner, which gives the network the flexibility to learn to ignore the random noise. In this paper, we introduce an approach to stochastically combine the root of variations with previous pose information, which forces the model to take the noise into account. We exploit this idea for motion prediction by incorporating it into a recurrent encoder-decoder network with a conditional variational autoencoder block that learns to exploit the perturbations. Our experiments demonstrate that our model yields high-quality pose sequences that are much more diverse than those from state-of-the-art stochastic motion prediction techniques. 

\end{abstract}

\section{Introduction}
Human motion prediction aims to forecast the sequence of future poses of a person given past observations of such poses. To achieve this, existing methods typically rely on recurrent neural networks (RNNs) that encode the person's motion~\cite{martinez2017human,gui2018adversarial,walker2017pose,kundu2018bihmp,barsoum2018hp,pavllo2019modeling,pavllo2018quaternet}. While they predict reasonable motions, RNNs are deterministic models and thus cannot account for the highly stochastic nature of human motion; given the beginning of a sequence, multiple, diverse futures are plausible. To correctly model this, it is therefore critical to develop algorithms that can learn the \emph{multiple modes} of human motion, even when presented with only deterministic training samples.

Recently, several attempts have been made at modeling the stochastic nature of human motion~\cite{yan2018mt,barsoum2018hp,walker2017pose,kundu2018bihmp,lin2018human}. These methods rely on sampling a random vector that is then combined with an encoding of the observed pose sequence. In essence, this combination is similar to the conditioning of generative networks; the resulting models aim to generate an output from a random vector while taking into account additional information about the content.

\begin{figure}
    \centering
    \small
        \input{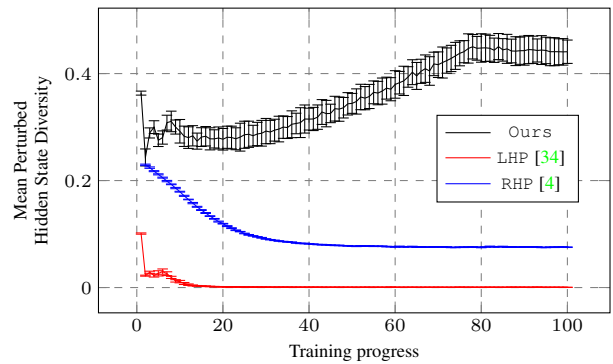} 
    \caption{Diversity of $K$ RNN decoder inputs,
    generated with $K=50$ different random vectors. We report the mean diversity over $N=50$ samples and the corresponding standard deviation. 
    }
    %\vspace{-10pt}
    \label{fig:baseline_RHP}
\end{figure}

Here, we argue that, while standard conditioning strategies may be
effective for many tasks, as in~\cite{yan2016attribute2image,kulkarni2015deep,esser2018variational,engel2017latent,bao2017cvae,larsen2015autoencoding}, they are ill-suited for motion prediction. The reason is the following: In other tasks, the conditioning variable only provides auxiliary information about the output to produce,
such as the fact that a generated face should be smiling. By contrast, in motion prediction, it typically contains the core signal to produce the output, i.e., the information about the previous poses. Since the prediction model is trained using deterministic samples, it can then simply learn to ignore the random vector and still produce a meaningful output based on the conditioning variable only. In other words, the model can ignore the root of variations, and thus essentially become deterministic. 

This problem was discussed in~\cite{bowman2015generating} in the context of text generation, and we identified it in our own motion prediction experiments. As evidence, we plot in Fig.~\ref{fig:baseline_RHP} the diversity of the representations used as input to the RNN decoders of~\cite{yan2018mt} (\texttt{LHP}) and~\cite{barsoum2018hp} (\texttt{RHP}), two state-of-the-art methods that are closest in spirit to our approach. Here, diversity is measured as the average pairwise distance across the $K$ representations produced for a single series of observations. We report the mean diversity over 50 samples and the corresponding standard deviation. As can be seen from the figure, the diversity of \texttt{LHP} and \texttt{RHP} decreases as training progresses, thus supporting our observation that the models learn to ignore the 
perturbations.

In this paper, we introduce a simple yet effective approach to counteracting this loss of diversity and thus to generating truly diverse future pose sequences. At the heart of our approach lies the idea of \emph{Mix-and-Match} perturbations: Instead of combining a noise vector with the conditioning variables in a deterministic manner, 
we randomly select and perturb a subset of these variables. By randomly changing this subset at every iteration, our strategy prevents training from identifying the root of variations and forces the model to take it into account in the generation process. As a consequence, and as evidenced by the black curve in Fig.~\ref{fig:baseline_RHP}, which shows an increasing diversity as training progresses, our approach produces not only high-quality predictions but also truly diverse ones.

In short, our contributions are \textbf{(1)} a novel way of imposing diversity into conditional VAEs, called \emph{Mix-and-Match perturbations}; \textbf{(2)} a new motion prediction model capable of generating multiple likely future pose sequences from an observed motion; and \textbf{(3)} a new evaluation metric for quantitatively measuring the quality and the diversity of generated motions, thus facilitating the comparison of different stochastic approaches.

\section{Related Work}
\label{sec:related_work}

Most motion prediction approaches are based on \emph{deterministic} models~\cite{pavllo2018quaternet,pavllo2019modeling,gui2018adversarial,jain2016structural,martinez2017human,gui2018few,fragkiadaki2015recurrent,ghosh2017learning}, casting motion prediction as a regression task where only one outcome is possible given the observations. While this may produce accurate predictions, it fails to reflect the stochastic nature of human motion, where multiple plausible outcomes can be highly likely for a single given series of observations. Modeling this diversity is the topic of this paper, and we therefore focus the discussion below on the other methods that have attempted to do so.

The general trend to incorporate variations in the predicted motions consists of combining information about the observed pose sequence with a random vector. In this context, two types of approaches have been studied: The techniques that directly incorporate the random vector into the RNN decoder and those that make use of an additional Conditional Variational Autoencoder (CVAE)~\cite{sohn2015learning}.

In the first class of methods,~\cite{lin2018human} samples a random vector $z_t\sim\mathcal{N}(0,I)$ at each time step and adds it to the pose input to the RNN decoder. By relying on different random vectors at each time step, however, this strategy is prone to generating discontinuous motions. To overcome this,~\cite{kundu2018bihmp} makes use of a single random vector to generate the entire sequence. This vector is both employed to alter the initialization of the decoder and concatenated with a pose embedding at each iteration of the RNN. By relying on concatenation, these two methods contain parameters that are specific to the random vector, and thus give the model the flexibility to ignore this information.
In~\cite{barsoum2018hp}, instead of using concatenation, the random vector is added to the hidden state produced by the RNN encoder. While addition prevents having parameters that are specific to the random vector, this vector is first transformed by multiplication with a parameter matrix, and thus can again be zeroed out so as to remove the source of diversity, as we observe empirically in Section~\ref{sec:experiments_QD}. In our experiments, we will refer to this method as \texttt{RHP}, for random hidden state perturbation.

The second category of stochastic methods introduce an additional CVAE between the RNN encoder and decoder. This allows them to learn a more meaningful transformation of the noise, combined with the conditioning variables, before passing the resulting information to the RNN decoder. In this context,~\cite{walker2017pose} proposes to directly use the pose as conditioning variable. As will be shown in our experiments, while this approach is able to maintain some degree of diversity, albeit less than ours, it yields motions of lower quality because of its use of independent random vectors at each time step. In our experiments, we will refer to this method as \texttt{LPP}, for learned pose perturbation.  In~\cite{butepage2018anticipating}, an approach similar to that of~\cite{walker2017pose} is proposed, but with one CVAE per limb. As such, this method suffers from the same discontinuity problem as~\cite{walker2017pose,lin2018human}. Finally, instead of perturbing the pose, the recent work~\cite{yan2018mt} uses the RNN decoder hidden state as conditioning variable in the CVAE, concatenating it with the random vector. While this approach generates high-quality motions, it suffers from the fact that the CVAE decoder gives the model the flexibility to ignore the random vector. In the remainder of the paper, we will refer to this method as \texttt{LHP}, for learned hidden state perturbation.

Ultimately, both classes of methods suffer from the fact that they allow the model to ignore the random vector, thus relying entirely on the conditioning information to generate future poses. Here, we introduce an effective way to maintain the root of diversity by randomizing the combination of the random vector with the conditioning variable.

\section{Proposed Method}
In this section, we first introduce our \emph{Mix-and-Match} approach to introducing diversity in CVAE-based motion prediction. We then describe the motion prediction architecture we used in our experiments and propose a novel evaluation metric to quantitatively measure the diversity and  quality of generated motions.

\begin{figure}
    \centering
    \includegraphics[width=0.5\textwidth]{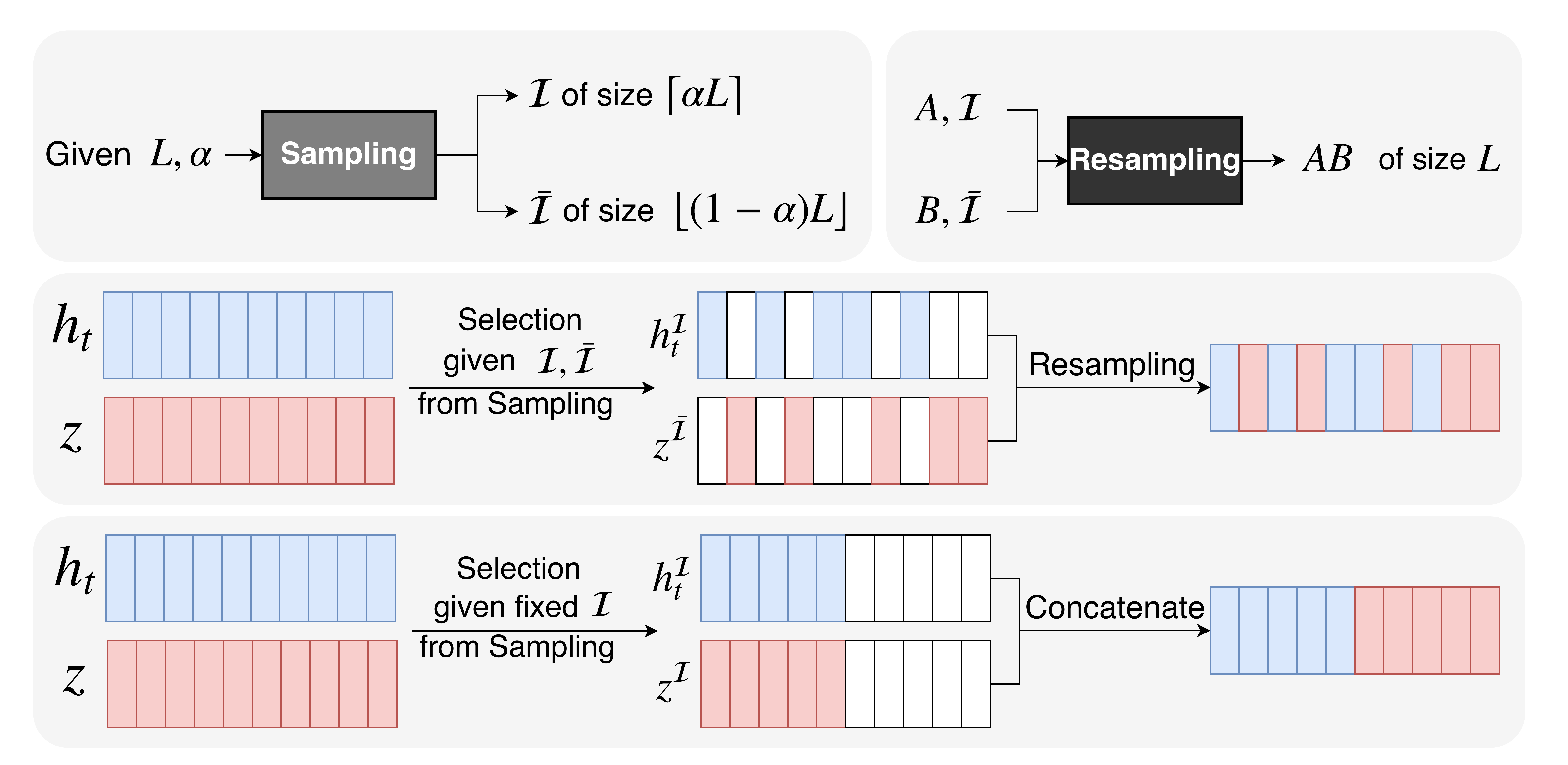}
    \caption{Mix-and-Match perturbation. \textbf{(Top)} Illustration of the \textit{Sampling} operation (left) and of the \textit{Resampling} one (right). Given a sampling rate $\alpha$ and a vector length $L$, the Sampling operation samples $\left\lceil\alpha L\right\rceil$ indices, say $\mathcal{I}$. The complementary, unsampled, indices are denoted by $\mathcal{\bar{I}}$. Then, given two $L$-dimensional vectors and the corresponding  $\left\lceil\alpha L\right\rceil$ and $\left\lfloor(1-\alpha)L\right\rfloor$ indices, the Resampling operation mixes the two vectors to form a new $L$-dimensional one. \textbf{(Middle)} Example of Mix-and-Match perturbation. \textbf{(Bottom)} Example of perturbation by concatenation, as in~\cite{yan2018mt}. Note that, in Mix-and-Match perturbations, sampling is stochastic; the indices are sampled uniformly randomly for each mini-batch. By contrast, in~\cite{yan2018mt}, sampling is deterministic, and the indices in $\mathcal{I}$ are fixed and correspond to $\mathcal{I}=\{1,\dots,\frac{L}{2}\}$.
    }
    \label{fig:mix_match}
\end{figure}

\subsection{Mix-and-Match Perturbation}
\label{sec:mix_match}

The main limitation of prior work in the area of stochastic motion modeling, such as~\cite{walker2017pose,barsoum2018hp,yan2018mt}, lies in the way they fuse the random vector with the conditioning variable, i.e., RNN hidden state or pose, which causes the model to learn to ignore the  randomness and solely exploit the deterministic conditioning information to generate motion~\cite{walker2017pose,barsoum2018hp,yan2018mt}. 
To overcome this, we propose to make it harder for the model to decouple the random variable from the deterministic information. Specifically, we observe that the way the random variable and the conditioning one are combined in existing methods is deterministic. We therefore propose to make this process stochastic.

Similarly to~\cite{yan2018mt}, we propose to make use of the hidden state as the conditioning variable and generate a perturbed hidden state by combining a part of the original hidden state with the random vector. However, as illustrated in Fig.~\ref{fig:mix_match}, instead of assigning predefined, deterministic indices to each piece of information, such as the first half for the hidden state and the second one for the random vector, we assign the values of hidden state to \textit{random} indices and the random vector to the complementary ones. 

More specifically, as depicted in Fig.~\ref{fig:mix_match}, a mix-and-match perturbation takes two vectors of size $L$ as input, say $h_t$ and $z$, and combines them in a stochastic manner. To this end, it relies on two operations. The first one, called \textit{Sampling}, chooses $\left\lceil\alpha L\right\rceil$ indices uniformly at random among the $L$ possible values, given a sampling rate $0\leq \alpha \leq 1$. Let us denote by ${\cal I} \subseteq\{1,\ldots,L\}$, the resulting set of indices and by ${\cal \bar{I}}$ the complementary set. The second operation, called \textit{Resampling}, then creates a new $L$-dimensional vector whose values at indices in $\mathcal{I}$ are taken as those at corresponding indices in the first input vector and the others at the complementary indices in the second input vector. Note that, the second vector can also have dimension $\left\lfloor(1-\alpha)L\right\rfloor$, and its values be divided among the remaining indices of the output vector.

\begin{figure*}[t]
    \centering
    \includegraphics[width=\textwidth]{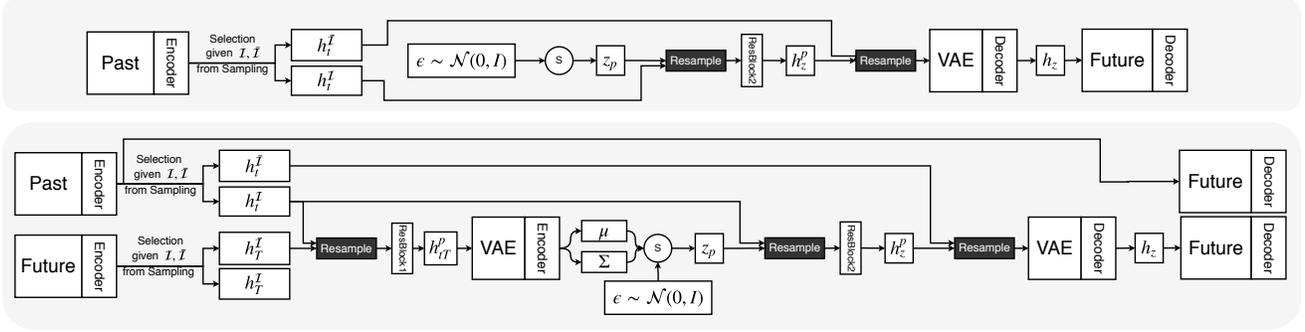}
    \caption{Overview of our approach. \textbf{(Top): Overview of the model during inference}. During inference, given past information and a random vector sampled from a Normal distribution, the model generate new motions.
    \textbf{(Bottom): Overview of the model during training.} During training, we use a future pose autoencoder with a CVAE  between the encoder and the decoder. The RNN encoder-decoder network mapping the past to the future then aims to generate good conditioning variables for the CVAE. }
    \label{fig:stochastic}
\end{figure*}

\subsection{M\&M Perturbation for Motion Prediction}
Let us now describe the way we use our mix-and-match perturbation strategy for motion prediction. To this end, we first discuss the network we rely on during inference, and then explain our training strategy.

\paragraph{Inference.} The high-level architecture we use at inference time is depicted by Fig.~\ref{fig:stochastic} (Top). It consists of an RNN encoder that takes $t$ poses $x_{1:t}$ as input and outputs an $L$-dimensional hidden vector $h_t$. A random $\left\lceil\alpha L\right\rceil$-dimensional portion of this hidden vector, $h_t^\mathcal{I}$, is then combined with an $\left\lfloor(1-\alpha)L\right\rfloor$-dimensional random vector $z\sim\mathcal{N}(0,I)$ via our mix-and-match perturbation strategy.
The resulting $L$-dimensional output is passed through a small neural network (i.e., \textit{ResBlock2} in Fig.~\ref{fig:stochastic}) that reduces its size to $\left\lceil\alpha L\right\rceil$, and then fused with the remaining $\left\lfloor(1-\alpha)L\right\rfloor$-dimensional portion of the hidden state, $h_t^\mathcal{\bar{I}}$. This, in turn, is passed through the VAE decoder to produce the final hidden state $h_z$, from which the future poses $x_{t+1:T}$ are obtained via the RNN decoder.

\paragraph{Training.} During training, we aim to learn both the RNN parameters and the CVAE ones. Because the CVAE is an \emph{auto}encoder, it needs to take as input information about future poses. To this end, we complement our inference architecture with an additional RNN future encoder, yielding the training architecture depicted in Fig.~\ref{fig:stochastic} (Bottom). Note that, in this architecture, we incorporate an additional mix-and-match perturbation that fuses the hidden state of the RNN past encoder $h_t$ with that of the RNN future encoder $h_T$ and forms $h_{tT}^p$. This allows us to condition the VAE encoder in a manner similar to the decoder.
Note that, for each mini batch, we use the same set of sampled indices for all mix-and-match perturbation steps throughout the network. Furthermore, following the standard CVAE strategy, during training, the random vector $z$ is sampled from a distribution $\mathcal{N}(\mu_\theta(x), \Sigma_\theta(x))$, whose mean $\mu_\theta(x)$ and covariance matrix $\Sigma_\theta(x)$ are produced by the CVAE encoder with parameters $\theta$. This is done by the technique of~\cite{kingma2013auto}, which computes $z_p$ as,

\begin{align}
z_p=\mu_\theta(x) + \Sigma_\theta(x)^{\frac{1}{2}} \epsilon\;,
\label{eq:reparameterization}
\end{align}
where $\epsilon\sim\mathcal{N}(0,I)$. Note that, during inference, $z_p=\epsilon$ since we do not have access to $x$, hence to $\mu_\theta(x)$ and $\Sigma_\theta(x)$.

To learn the parameters of our model, we rely on the availability of a dataset $D=\{X_1, X_2, ..., X_N\}$ containing $N$ videos $X_i$ depicting a human performing an action. Each video consists of a sequence of $T$ poses, $X_i=\{x_i^1, x_i^2, ..., x_i^T\}$, and each pose comprises $J$ joints forming a skeleton, $x_i^t=\{x_{i, 1}^t, x_{i, 2}^t, ..., x_{i, J}^t\}$. The pose of each joint is represented as a quaternion. Given this data, we train our model by minimizing a loss function of the form
\begin{align}
\mathcal{L} = \frac{1}{N} \sum_{i=1}^N\Big(\mathcal{L}_{rot}(X_i) + \mathcal{L}_{skl}(X_i)\Big) + \lambda\mathcal{L}_{prior}\;.
\label{eq:stochastic_loss}
\end{align}

The first term in this loss compares the output of the network with the ground-truth motion using the squared loss. That is, 
\begin{align}
\mathcal{L}_{rot}(X_i) = -\frac{1}{T-t}\frac{1}{4J}\sum_{k=t+1}^T\sum_{j=1}^{J}{\|\hat{x}^{k}_{i,j} - x^{k}_{i,j}\|}^2\;,
\label{eq:loss_quat}
\end{align}
where $\hat{x}^{k}_{i,j}$ is the predicted 4D quaternion for the $j^{th}$ joint at time $k$ in sample $i$, and $x^{k}_{i,j}$ the corresponding ground-truth one. The main weakness of this loss is that it treats all joints equally. However, when working with angles, some joints have a much larger influence on the pose than others. For example, because of the kinematic chain, the pose of the shoulder affects that of the rest of the arm, whereas the pose of the wrists has only a minor effect.

To take this into account, we define our second loss term as the error in 3D space. That is, 
\begin{align}
\mathcal{L}_{skl}(X_i) = -\frac{1}{T-t}\frac{1}{J\times 3}\sum_{k=t+1}^{T}\sum_{j=1}^{J} \|\hat{p}^{k}_{i,j} - p^{k}_{i,j}\|^2\;,
\label{eq:loss_skl}
\end{align}
where $\hat{p}^{k}_{i,j}$ is the predicted 3D position of joint $j$ at time $k$ in sample $i$ and $p^{k}_{i,j}$ the corresponding ground-truth one. These 3D positions can be computed using forward kinematics, as in~\cite{pavllo2018quaternet,pavllo2019modeling}. Note that, to compute this loss, we first perform a global alignment of the predicted pose and the ground-truth one by rotating the root joint to face [0, 0, 0].

Finally, following standard practice in training VAEs, we define our third loss term as the KL divergence 
\begin{align}
    \mathcal{L}_{prior} = -KL\Big(\mathcal{N}(\mu_\theta(x), \Sigma_\theta(x))) \Big\| \mathcal{N}(0,I)\Big)\;.
    \label{eq:kl}
\end{align}

In practice, since our VAE appears within a recurrent model, we weigh $\mathcal{L}_{prior}$ by a function $\lambda$ corresponding to the KL annealing weight of~\cite{bowman2015generating}. We start from $\lambda=0$, forcing the model to encode as much information in $z$ as possible, and gradually increase it to $\lambda=1$, following a logistic curve.

\paragraph{Curriculum Learning of Variation.} The parameter $\alpha$ in our mix-and-match perturbation scheme determines a trade-off between stochasticity and motion quality. The larger $\alpha$, the larger the portion of the original hidden state that will be perturbed. Thus, the model incorporates more randomness and less information from the original hidden state. As such, given a large $\alpha$, it becomes harder for the model to deliver motion information from the observation to the future representation since a large portion of the hidden state is changing randomly. In particular, we observed that training becomes unstable if we use a large $\alpha$ from the beginning, with the motion-related loss terms fluctuating while the prior loss $\mathcal{L}_{prior}$ quickly converges to zero. 

\begin{figure}
    \centering
    \includegraphics[width=0.45\textwidth]{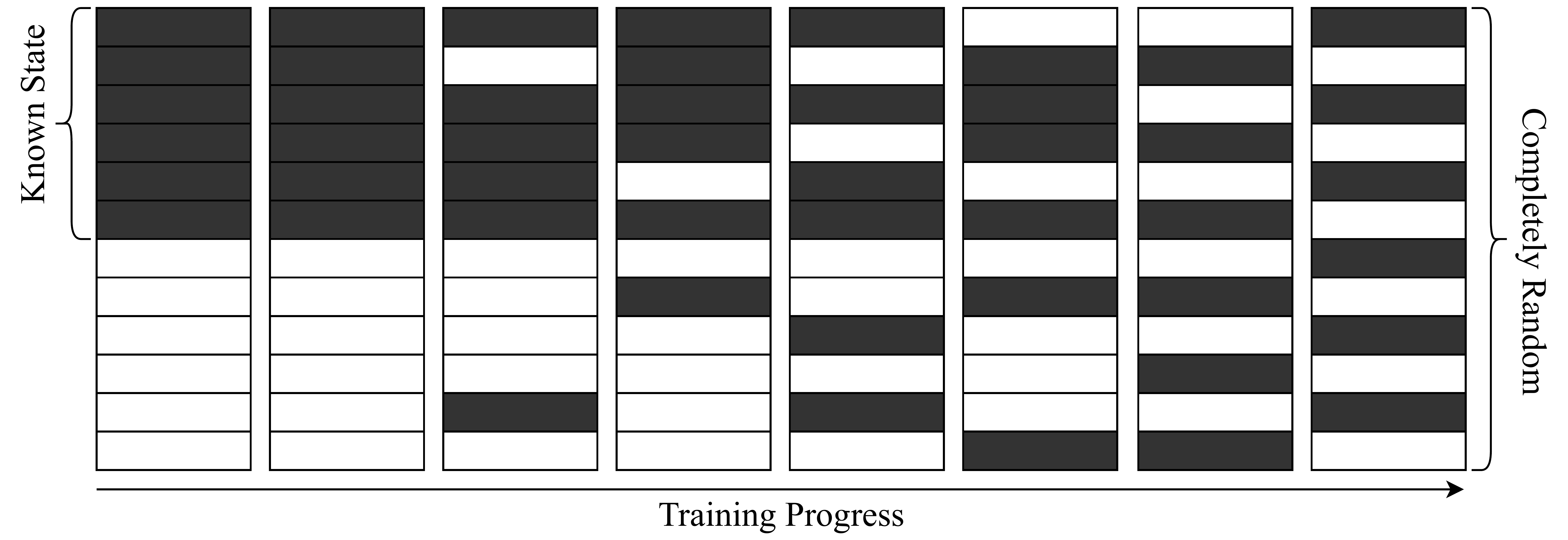}
    \caption{Example of curriculum perturbation of the hidden state. At the beginning of training, the perturbation occurs in a deterministic portion of the hidden state. As training progresses, we gradually, and randomly, spread the perturbation to the rest of the hidden state. This continues until the indices to perturb are uniformly randomly sampled.
    }
    \label{fig:curriculum}
\end{figure}

To overcome this while still enabling the use of sufficiently large values of $\alpha$ to achieve high diversity, we introduce the curriculum learning strategy depicted by Fig.~\ref{fig:curriculum}. In essence, we initially select $\left\lceil\alpha L\right\rceil$ indices in a deterministic manner and gradually increase the randomness of these indices as training progresses. More specifically, given a set of $\left\lceil\alpha L\right\rceil$ indices, we replace $c$ indices from the sampled ones with the corresponding ones from the remaining $\left\lfloor(1-\alpha)L\right\rfloor$ indices. Starting from $c=0$, we gradually increase $c$ to the point where all $\left\lceil\alpha L\right\rceil$ indices are sampled uniformly randomly. More details, including the pseudo-code of this approach, are provided in the supplementary material. This strategy helps the motion decoder to initially learn and incorporate information about the observations (as in~\cite{yan2018mt}), yet, in the long run, still prevents it from ignoring the random vector.

\subsection{Quality and Diversity Metrics}
\label{sec:eval}
When dealing with multiple plausible motions, or in general diverse solutions to a problem, evaluation is a challenge. The standard metrics used for deterministic motion prediction models are ill-suited to this task, because they typically compare the predictions to the ground truth, thus inherently penalizing diversity. For multiple motions, two aspects are important: the \emph{diversity} and the \emph{quality}, or realism, of each individual motion.
Prior work typically evaluates these aspects via human judgement. While human evaluation is highly valuable, and we will also report human results, it is very costly and time-consuming. Here, we therefore introduce two metrics that facilitate the quantitative evaluation of both quality and diversity.
 
To measure the quality of generated motions, we propose to rely on a binary classifier 
trained to discriminate real (ground-truth) samples from fake (generated) ones. The accuracy of this classifier on the test set is thus inversely proportional to the quality of the generated motions. In other words, high-quality motions are those that are not distinguishable from real ones. Note that we do not advocate for adversarial training of our approach. That is, we do not define a loss based on this classifier when training our model.
 
To measure the diversity of the generated motions, a naive approach would consist of relying on the distance between the generated motion and a reference one. However, generating identical motions that are all far from the reference one would therefore yield a high value, while not reflecting diversity. To prevent this, we therefore propose to make use of the average distance between all pairs of generated motions.

\section{Experiments}

Let us now evaluate the effectiveness of our approach at generating multiple plausible motions. To this end, we use Human3.6M~\cite{h36m_pami}, the largest publicly available motion capture dataset.  Below, we first exploit the metrics introduced in Section~\ref{sec:eval} to compare the quality and diversity of the results of our approach with those obtained by state-of-the-art methods that produce multiple motions~\cite{yan2018mt, walker2017pose,barsoum2018hp}. We then compare our results to the state-of-the-art deterministic motion prediction techniques for long-term motion prediction using standard metrics.

\paragraph{Implementation details.} The motion encoders and decoders in our model are single layer GRU~\cite{cho2014learning} networks, comprising 1024 hidden units each. For the decoders, we use a teacher forcing technique~\cite{williams1989learning} to decode motion. At each time-step, the network chooses with probability $P_{tf}$ whether to use its own output at the previous time-step or the ground-truth pose as input. We initialize $P_{tf}=1$, and decrease it linearly at each training epoch such that, after a certain number of epochs, the model becomes completely autoregressive, i.e., uses only its own output as input to the next time-step. 
We train our model on a single GPU with the Adam optimizer~\cite{kingma2014adam} for 100K iterations. We use a learning rate of 0.001 and a mini-batch size of 64. To avoid exploding gradients, we use the gradient-clipping technique of~\cite{pascanu2013difficulty} for all layers in the network. We implemented our model using the Pytorch framework of~\cite{paszke2017automatic}.

\subsection{Evaluating Quality and Diversity}
\label{sec:experiments_QD}
Quantitative evaluation of a qualitative task is very challenging. While the ideal case is reporting the (log-)likelihood on a held-out set of samples, in (non-probabilistic) decoder-based generative models this is not possible. An alternative is using non-parametric kernel density estimates (KDE), only via samples, however, KDE is only well-suited for very low dimensional data space. 
Evaluating against one GT motion (i.e., one sample from multi-modal distribution) can lead to a high score for one sample while penalizing other plausible modes. This behaviour is undesirable since it cannot differentiate a multi-modal solution with a good, but uni-modal one. Note, there exist some metrics~\cite{yan2018mt} to evaluate motions, however, they do not reflect the quality of a prediction, but how likely ground-truth future motions are with the given model. Moreover, as discussed, the metrics in~\cite{yan2018mt} only evaluate quality given one single groundtruth. While the groundtruth has high quality, there exist multiple high quality continuations of an observation, which our metric accounts for.
As discussed in Section~\ref{sec:eval}, we evaluate both the quality and diversity of the predicted motions. Note, these two metrics should be considered together, since each one taken separately does not provide a complete picture of how well a model can predict \emph{multiple plausible} future motions. For example, a model can generate diverse but unnatural motions, or, conversely, realistic but identical motions.

We compare our \emph{Mix-and-Match} approach with the different means of imposing variation in motion prediction discussed in Section~\ref{sec:related_work}, i.e., concatenating the hidden state to a learned latent variable (\texttt{LHP})~\cite{yan2018mt}, concatenating the pose to a learned latent variable at each time-step (\texttt{LPP})~\cite{walker2017pose}, and adding a (transformed) random noise to the hidden state (\texttt{RHP})~\cite{barsoum2018hp}. For the comparison to be fair, we use 16 frames (i.e., 640ms) as observation to generate the next 60 frames (i.e., 2.4sec) for all baselines. All models are trained with the same motion representation, backbone network, and  losses, except for \texttt{RHP} which cannot make use of $\mathcal{L}_{prior}$.

\begin{figure}
    \centering
    \includegraphics[width=.4\textwidth]{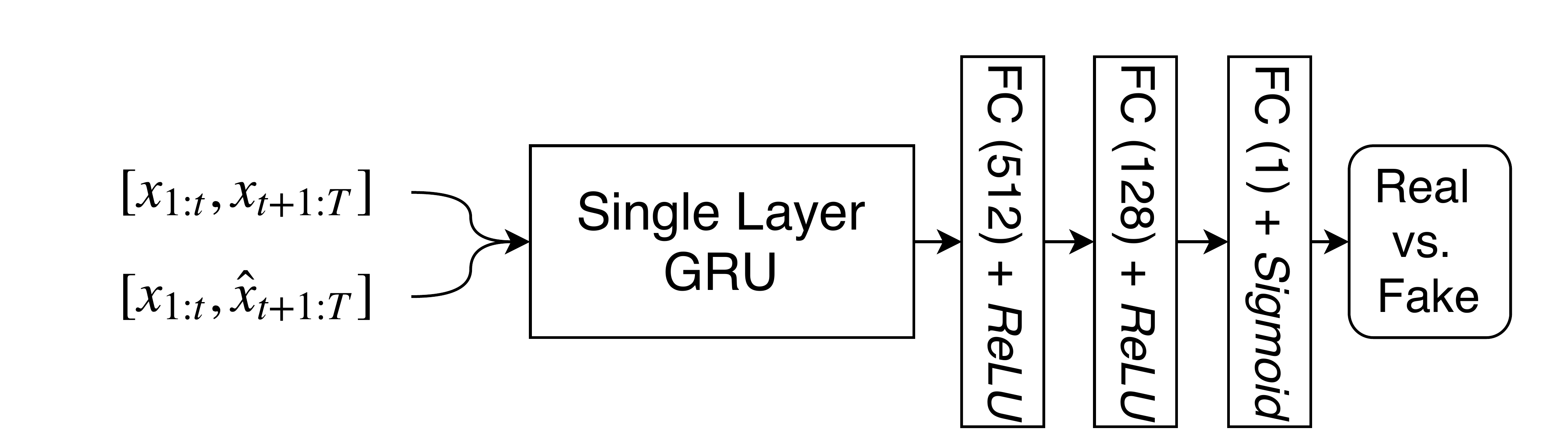}
    \caption{Architecture of the quality binary classifier.}
    \label{fig:quality_classifier}
\end{figure}

\begin{figure}
    \centering
    \begin{tabular}{c}
    \pgfplotsset{width=0.48\textwidth, height=0.2\textheight, compat=1.3, legend columns=2,  legend style={at={(0.7,0.97)},anchor=north}, grid style={dashed,gray}}    
\begin{tikzpicture}
\scriptsize
\begin{axis}[
%   axis y line*=left,
%   axis x line=none,
% xlabel=Training progress,
  ymin=0, ymax=6,
  ylabel={\scriptsize Diversity},
  %style={thick},
 grid=both
]

\addplot[smooth,red, style={thick}]
  coordinates{
(	2	,	0.372228319	)
(	4	,	1.3713228096	)
(	6	,	0.9580214302	)
(	8	,	0.8025371043	)
(	10	,	0.7632041583	)
(	12	,	0.7101821605	)
(	14	,	0.7091121015	)
(	16	,	0.6972464916	)
(	18	,	0.6879635837	)
(	20	,	0.6569117425	)
(	22	,	0.6385426795	)
(	24	,	0.6134333652	)
(	26	,	0.6368943344	)
(	28	,	0.619573475	)
(	30	,	0.6237382963	)
(	32	,	0.6008411544	)
(	34	,	0.6032496738	)
(	36	,	0.5877619199	)
(	38	,	0.5884365005	)
(	40	,	0.580042939	)
(	42	,	0.5703011194	)
(	44	,	0.5657127558	)
(	46	,	0.5640289707	)
(	48	,	0.5516945977	)
(	50	,	0.5468479843	)
(	52	,	0.5494695269	)
(	54	,	0.5531247495	)
(	56	,	0.5417309554	)
(	58	,	0.5289380869	)
(	60	,	0.5297903089	)
(	62	,	0.533012786	)
(	64	,	0.5160716335	)
(	66	,	0.5182542383	)
(	68	,	0.5083657478	)
(	70	,	0.5158736547	)
(	72	,	0.5141236558	)
(	74	,	0.4938643058	)
(	76	,	0.5027989279	)
(	78	,	0.508373203	)
(	80	,	0.5012296765	)
(	82	,	0.4878586102	)
(	84	,	0.4786059192	)
(	86	,	0.4860810223	)
(	88	,	0.4903411442	)
(	90	,	0.4895399712	)
(	92	,	0.4835897147	)
(	94	,	0.4816715511	)
(	96	,	0.4795188333	)
(	98	,	0.4856411611	)
(	100	,	0.4884118867	)

}; \addlegendentry{\texttt{RHP}}

\addplot[smooth,blue, style={thick}]
  coordinates{
(	2	,	2.1994370685	)
(	4	,	5.6279087043	)
(	6	,	3.8467925915	)
(	8	,	2.9263108718	)
(	10	,	2.2151651559	)
(	12	,	2.0074867801	)
(	14	,	1.6942524325	)
(	16	,	1.5318150884	)
(	18	,	1.4325693004	)
(	20	,	1.3508969624	)
(	22	,	1.3865340326	)
(	24	,	1.2542548549	)
(	26	,	1.2954020445	)
(	28	,	1.184913049	)
(	30	,	1.2392374261	)
(	32	,	1.2847838287	)
(	34	,	1.2748868949	)
(	36	,	1.2639755715	)
(	38	,	1.283692168	)
(	40	,	1.2876274793	)
(	42	,	1.3695187998	)
(	44	,	1.3514408685	)
(	46	,	1.41878608	)
(	48	,	1.428349829	)
(	50	,	1.434303543	)
(	52	,	1.4803239765	)
(	54	,	1.516733712	)
(	56	,	1.5193682853	)
(	58	,	1.5805738382	)
(	60	,	1.5893576212	)
(	62	,	1.6105508847	)
(	64	,	1.6157088854	)
(	66	,	1.686402194	)
(	68	,	1.6968436156	)
(	70	,	1.6913643432	)
(	72	,	1.6970748497	)
(	74	,	1.7221260986	)
(	76	,	1.740707242	)
(	78	,	1.7148590457	)
(	80	,	1.748023004	)
(	82	,	1.7127991344	)
(	84	,	1.7537020766	)
(	86	,	1.7332248949	)
(	88	,	1.7164786857	)
(	90	,	1.7581485555	)
(	92	,	1.7356540302	)
(	94	,	1.7248771592	)
(	96	,	1.7199494125	)
(	98	,	1.7191962364	)
(	100	,	1.7055290809	)

}; \addlegendentry{\texttt{LPP}}

\addplot[smooth,orange, style={thick}]
  coordinates{
(	2	,	2.138264193	)
(	4	,	0.9583889843	)
(	6	,	0.9864167158	)
(	8	,	1.432723681	)
(	10	,	1.196091712	)
(	12	,	1.599456923	)
(	14	,	1.676673717	)
(	16	,	1.849031882	)
(	18	,	1.462363488	)
(	20	,	1.137150942	)
(	22	,	0.7486108682	)
(	24	,	0.7131989401	)
(	26	,	0.3123724991	)
(	28	,	0.2893492722	)
(	30	,	0.2827597115	)
(	32	,	0.2808223394	)
(	34	,	0.238696808	)
(	36	,	0.2912405132	)
(	38	,	0.2405678643	)
(	40	,	0.2423426698	)
(	42	,	0.2529671774	)
(	44	,	0.2450041449	)
(	46	,	0.225672877	)
(	48	,	0.2692853561	)
(	50	,	0.217450544	)
(	52	,	0.2357198347	)
(	54	,	0.3292924177	)
(	56	,	0.2485134083	)
(	58	,	0.2256134417	)
(	60	,	0.475737121	)
(	62	,	0.2523357689	)
(	64	,	0.2632740174	)
(	66	,	0.2661004258	)
(	68	,	0.302793266	)
(	70	,	0.285583902	)
(	72	,	0.2416136429	)
(	74	,	0.2446458042	)
(	76	,	0.4140788055	)
(	78	,	0.3244524419	)
(	80	,	0.2605087185	)
(	82	,	0.2692853561	)
(	84	,	0.217450544	)
(	86	,	0.2357198347	)
(	88	,	0.3292924177	)
(	90	,	0.2485134083	)
(	92	,	0.302793266	)
(	94	,	0.335737121	)
(	96	,	0.285583902	)
(	98	,	0.2632740174	)
(	100	,	0.2661004258	)

}; \addlegendentry{\texttt{LHP}}

\addplot[smooth,black, style={thick}]
  coordinates{
(	2	,	0.2409195131	)
(	4	,	1.6280298248	)
(	6	,	2.0981391871	)
(	8	,	2.1861691242	)
(	10	,	2.0288539579	)
(	12	,	1.9899903938	)
(	14	,	2.1445177616	)
(	16	,	2.1318833053	)
(	18	,	2.292951828	)
(	20	,	2.3463853261	)
(	22	,	2.4200368508	)
(	24	,	2.570214483	)
(	26	,	2.6435528647	)
(	28	,	2.698081253	)
(	30	,	2.7387426613	)
(	32	,	2.7749708101	)
(	34	,	2.8341599759	)
(	36	,	2.8538404449	)
(	38	,	2.9834643572	)
(	40	,	2.9553836636	)
(	42	,	2.9727804663	)
(	44	,	3.1151485337	)
(	46	,	3.1040881512	)
(	48	,	3.0891438455	)
(	50	,	3.1701705792	)
(	52	,	3.1011047392	)
(	54	,	3.1829042973	)
(	56	,	3.2342386935	)
(	58	,	3.2590998386	)
(	60	,	3.2688277237	)
(	62	,	3.3114212448	)
(	64	,	3.3906127828	)
(	66	,	3.3778737181	)
(	68	,	3.4632934095	)
(	70	,	3.497815927	)
(	72	,	3.531132392	)
(	74	,	3.5635725847	)
(	76	,	3.5829808273	)
(	78	,	3.6399284083	)
(	80	,	3.6326588548	)
(	82	,	3.5910308341	)
(	84	,	3.5782219032	)
(	86	,	3.5734242093	)
(	88	,	3.5775188958	)
(	90	,	3.5347458896	)
(	92	,	3.5816775293	)
(	94	,	3.5099271611	)
(	96	,	3.5179532774	)
(	98	,	3.4796801437	)
(	100	,	3.5206109798	)

}; \addlegendentry{\texttt{Ours}}

\end{axis}

\end{tikzpicture} \\
    \pgfplotsset{width=0.48\textwidth,height=0.2\textheight, compat=1.3, legend columns=2,  legend style={at={(0.7,.7)},anchor=north}, grid style={dashed,gray}}    
\begin{tikzpicture}
\scriptsize
\begin{axis}[
%   axis y line*=left,
  ymin=0, ymax=50,
   xlabel=Training progress,
  ylabel={\scriptsize Quality},
  grid=both
%   style={thick}
]
\addplot[smooth,orange, style={thick}]
  coordinates{
(2,	6)
(4,	13)
(6,	20)
(8,	24)
(10,	29)
(12,	31)
(14,	33)
(16,	35)
(18,	36)
(20,	37)
(22,	39)
(24,	41)
(26,	43)
(28,	44)
(30,	45)
(32,	45)
(34,	45)
(36,	45)
(38,	45)
(40,	46)
(42,	46)
(44,	45)
(46,	45)
(48,	45)
(50,	46)
(52,	45)
(54,	45)
(56,	45)
(58,	45)
(60,	45)
(62,	45)
(64,	45)
(66,	45)
(68,	45)
(70,	44)
(72,	45)
(74,	45)
(76,	45)
(78,	45)
(80,	45)
(82,	45)
(84,	45)
(86,	44)
(88,	45)
(90,	46)
(92,	45)
(94,	46)
(96,	45)
(98,	45)
(100,	45)

}; \label{Hidden Quality}
\addlegendentry{\texttt{LHP}}

\addplot[smooth,blue, style={thick}]
  coordinates{
(	2	,	0	)
(	4	,	0	)
(	6	,	0	)
(	8	,	1	)
(	10	,	3	)
(	12	,	6	)
(	14	,	8	)
(	16	,	9	)
(	18	,	10	)
(	20	,	11	)
(	22	,	12	)
(	24	,	13	)
(	26	,	12	)
(	28	,	12	)
(	30	,	12	)
(	32	,	13	)
(	34	,	12	)
(	36	,	12	)
(	38	,	13	)
(	40	,	12	)
(	42	,	12	)
(	44	,	12	)
(	46	,	13	)
(	48	,	12	)
(	50	,	14	)
(	52	,	13	)
(	54	,	13	)
(	56	,	12	)
(	58	,	10	)
(	60	,	11	)
(	62	,	12	)
(	64	,	11	)
(	66	,	11	)
(	68	,	13	)
(	70	,	12	)
(	72	,	12	)
(	74	,	13	)
(	76	,	12	)
(	78	,	13	)
(	80	,	14	)
(	82	,	15	)
(	84	,	14	)
(	86	,	14	)
(	88	,	14	)
(	90	,	14	)
(	92	,	15	)
(	94	,	14	)
(	96	,	15	)
(	98	,	14	)
(	100	,	13	)

}; \label{Pose Noise Quality}
\addlegendentry{\texttt{LPP}}

\addplot[smooth,red, style={thick}]
  coordinates{
(	2	,	5	)
(	4	,	15	)
(	6	,	24	)
(	8	,	24	)
(	10	,	27	)
(	12	,	30	)
(	14	,	32	)
(	16	,	35	)
(	18	,	37	)
(	20	,	40	)
(	22	,	40	)
(	24	,	42	)
(	26	,	42	)
(	28	,	43	)
(	30	,	43	)
(	32	,	42	)
(	34	,	44	)
(	36	,	44	)
(	38	,	43	)
(	40	,	45	)
(	42	,	42	)
(	44	,	44	)
(	46	,	44	)
(	48	,	45	)
(	50	,	44	)
(	52	,	45	)
(	54	,	45	)
(	56	,	45	)
(	58	,	45	)
(	60	,	44	)
(	62	,	45	)
(	64	,	45	)
(	66	,	45	)
(	68	,	45	)
(	70	,	45	)
(	72	,	46	)
(	74	,	45	)
(	76	,	46	)
(	78	,	46	)
(	80	,	45	)
(	82	,	46	)
(	84	,	46	)
(	86	,	47	)
(	88	,	47	)
(	90	,	46	)
(	92	,	45	)
(	94	,	46	)
(	96	,	46	)
(	98	,	47	)
(	100	,	47	)

}; \label{White Noise Quality}
\addlegendentry{\texttt{RHP}}

\addplot[smooth,black, style={thick}]
  coordinates{
(	2	,	5	)
(	4	,	7	)
(	6	,	14	)
(	8	,	18	)
(	10	,	22	)
(	12	,	23	)
(	14	,	25	)
(	16	,	25	)
(	18	,	26	)
(	20	,	27	)
(	22	,	27	)
(	24	,	28	)
(	26	,	28	)
(	28	,	29	)
(	30	,	30	)
(	32	,	30	)
(	34	,	31	)
(	36	,	32	)
(	38	,	32	)
(	40	,	34	)
(	42	,	35	)
(	44	,	37	)
(	46	,	37	)
(	48	,	38	)
(	50	,	39	)
(	52	,	38	)
(	54	,	39	)
(	56	,	40	)
(	58	,	40	)
(	60	,	40	)
(	62	,	40	)
(	64	,	41	)
(	66	,	40	)
(	68	,	41	)
(	70	,	41	)
(	72	,	42	)
(	74	,	42	)
(	76	,	42	)
(	78	,	42	)
(	80	,	42	)
(	82	,	43	)
(	84	,	43	)
(	86	,	42	)
(	88	,	42	)
(	90	,	42	)
(	92	,	42	)
(	94	,	43	)
(	96	,	42	)
(	98	,	42	)
(	100	,	42	)

}; \label{Ours Quality}
\addlegendentry{\texttt{Ours}}

\end{axis}

\end{tikzpicture}
             
    \end{tabular}
    \caption{Quality and diversity evaluation. Our approach outperforms the baselines in terms of diversity while preserving a high quality, especially late in the training progress.}
    \label{fig:QD_plot}
\end{figure}
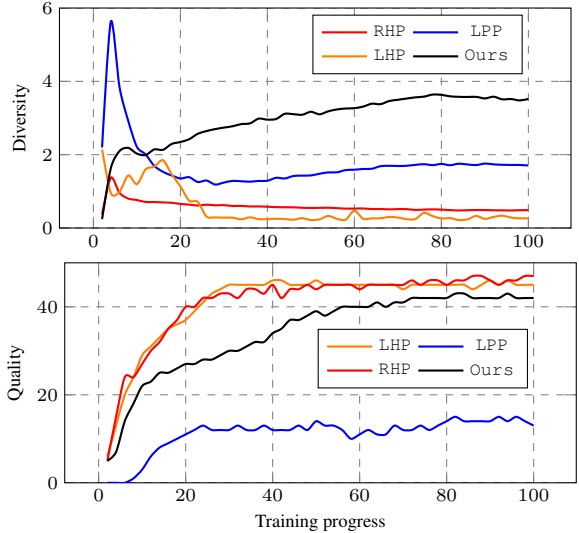

To evaluate quality, as discussed in Section~\ref{sec:eval}, we use a recurrent binary classifier whose task is to determine whether a sample comes from the ground-truth data or was generated by the model.
As depicted by Fig.~\ref{fig:quality_classifier}, the model is based on a single layer GRU network with 1024 hidden units to process the motion, followed by a three-layer fully connected network (with 512, 128 and 1 units, respectively) with ReLU non-linearity in between and a sigmoid non-linearity for binary classification. We train such a classifier for each method, using 25K samples generated at different training steps together with 25K real samples, forming a binary dataset of 50K motions for each method. We use stochastic gradient descent for 5K iterations, with a mini-batch size of 256, a learning rate of 0.01 and a momentum of 0.9. To evaluate diversity, as discussed in Section~\ref{sec:eval}, we compute the mean Euclidean distance from each motion to all other $K-1$ motions when generating $K=50$ motions. Furthermore, we also performed a human evaluation to measure the quality of the motions generated by each method. To this end, we asked eight users to rate the quality of 50 motions generated by each method, for a total of 200 motions. The ratings were defined on a scale of 1-5, 1 representing a low-quality motion and 5 a high-quality, realistic one. We then scaled the values to the range 0-50 to make them comparable with those of the binary classifier.

\begin{figure}
    \centering
    \pgfplotsset{width=0.4\textwidth, height=0.22\textheight, compat=1.3, legend columns=1,  legend style={at={(1.2,1)},anchor=north}, grid style={dashed,gray}}  
\usetikzlibrary{plotmarks}
\usetikzlibrary{backgrounds}
\begin{tikzpicture}

\begin{axis}[
%   axis y line*=left,
  ymin=0, ymax=4,
  xmin=0, xmax=50,
   xlabel={\scriptsize Quality},
  ylabel={\scriptsize Diversity},
  grid=both
]
\scriptsize
% \begin{scope}[on background layer]
    
%     \fill[green,opacity=0.25]  ({rel axis cs:0.5,0.5}) rectangle ({rel axis cs:1,1});
%     \fill[red,opacity=0.25]  ({rel axis cs:0,0}) rectangle ({rel axis cs:0.5,0.5});
    
%     \fill[orange,opacity=0.25]  ({rel axis cs:0,1}) rectangle ({rel axis cs:0.5,0.5});
%     \fill[orange,opacity=0.25]  ({rel axis cs:1,0}) rectangle ({rel axis cs:0.5,0.5});
    
%     % \fill[orange,opacity=0.5]  ({rel axis cs:0,0.6}) rectangle ({rel axis cs:1,.4});
%     % \fill[orange,opacity=0.5]  ({rel axis cs:0.4,0}) rectangle ({rel axis cs:0.6,1});
    
% \end{scope}

\addplot[mark=otimes*, mark size=3pt, blue]
  coordinates{
(50,	0)
}; \label{Real Motion}
\addlegendentry{\tiny Real Motion}

\addplot[mark=cube*, mark size=3pt, gray]
  coordinates{
(42,	3.52)
}; \label{Ours Classifier}
\addlegendentry{\tiny \texttt{Ours} (C)}

\addplot[mark=cube*, mark size=3pt, black]
  coordinates{
(40.92,	3.52)
}; \label{Ours Human}
\addlegendentry{\tiny \texttt{Ours} (H)}

\addplot[mark=pentagon*, mark size=3pt, gray]
  coordinates{
(45,	0.26)
}; \label{ECCV Classifier}
\addlegendentry{\tiny \texttt{LHP} (C)}
\addplot[mark=pentagon*, mark size=3pt, black]
  coordinates{
(37.0,	0.26)
}; \label{Ours Human}
\addlegendentry{\tiny \texttt{LHP} (H)}

\addplot[mark=otimes*, mark size=3pt, gray]
  coordinates{
(47,	0.48)
}; \label{CVPR Classifier}
\addlegendentry{\tiny \texttt{RHP} (C)}
\addplot[mark=otimes*, mark size=3pt, black]
  coordinates{
(38.12,	0.48)
}; \label{CVPR Human}
\addlegendentry{\tiny \texttt{RHP} (H)}

\addplot[mark=triangle*, mark size=3pt, gray]
  coordinates{
(13,	1.7)
}; \label{ICCV Classifier}
\addlegendentry{\tiny \texttt{LPP} (C)}
\addplot[mark=triangle*, mark size=3pt, black]
  coordinates{
(15,	1.7)
}; \label{ICCV Human}
\addlegendentry{\tiny \texttt{LPP} (H)}

% \addplot[dashed, black, style={thick}]
% coordinates{(0,2) (50,2)};
% \addplot[dashed, black, style={thick}]
% coordinates{(25,0) (25,4)};

\end{axis}
\end{tikzpicture}
    \caption{Human (H) and classifier-based (C) evaluation of quality for different methods. We plot diversity vs quality. A good method should fall into the top-right part of the plot, i.e., have high quality and diversity. Only our approach, for both human and classifier-based evaluation, satisfies this criterion. 
    Real motions (blue circle) are deterministic, i.e., one future per observation, and have 0 diversity. However, their quality is optimal, i.e., 50\%.
    Note that the human and classifier-based results depict the same behavior.
    }
    \label{fig:human_eval}
\end{figure}
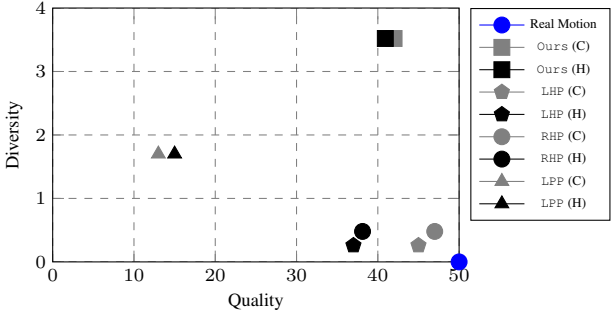

The results of the metrics of Section~\ref{sec:eval} are provided in  Fig.~\ref{fig:QD_plot} and those of the human evaluation in Fig.~\ref{fig:human_eval}. Below, we analyze the results of the different models.

\paragraph{\texttt{LHP}.} As can be seen from Fig.~\ref{fig:QD_plot}, \texttt{LHP} tends to ignore the random variable $z$, thus ignoring the root of variation. As a consequence, it achieves a low diversity, much lower than ours, but produces samples of high quality, albeit almost identical. Note that this decrease in diversity occurs after 16K iterations, indicating that the model takes time to identify the part of the hidden state that contains the randomness. Nevertheless, at iteration 16K, prediction quality is low, and thus one could not simply stop training at this stage. Note that the lack of diversity of \texttt{LHP} is also evidenced by Fig.~\ref{fig:baseline_RHP}. To further confirm it, we performed an additional experiment where, at test time, we sampled each element of the random vector independently from $\mathcal{N}(50, 50)$ instead of from the prior $\mathcal{N}(0, I)$. This led to neither loss of quality nor increase of diversity of the generated motions. As can be verified in Fig.~\ref{fig:human_eval}, where \texttt{LHP} appears in a region of high quality but low diversity, the results of human evaluation match those of our classifier-based quality metric.

\begin{figure*}
    \centering
    \includegraphics[width=\textwidth]{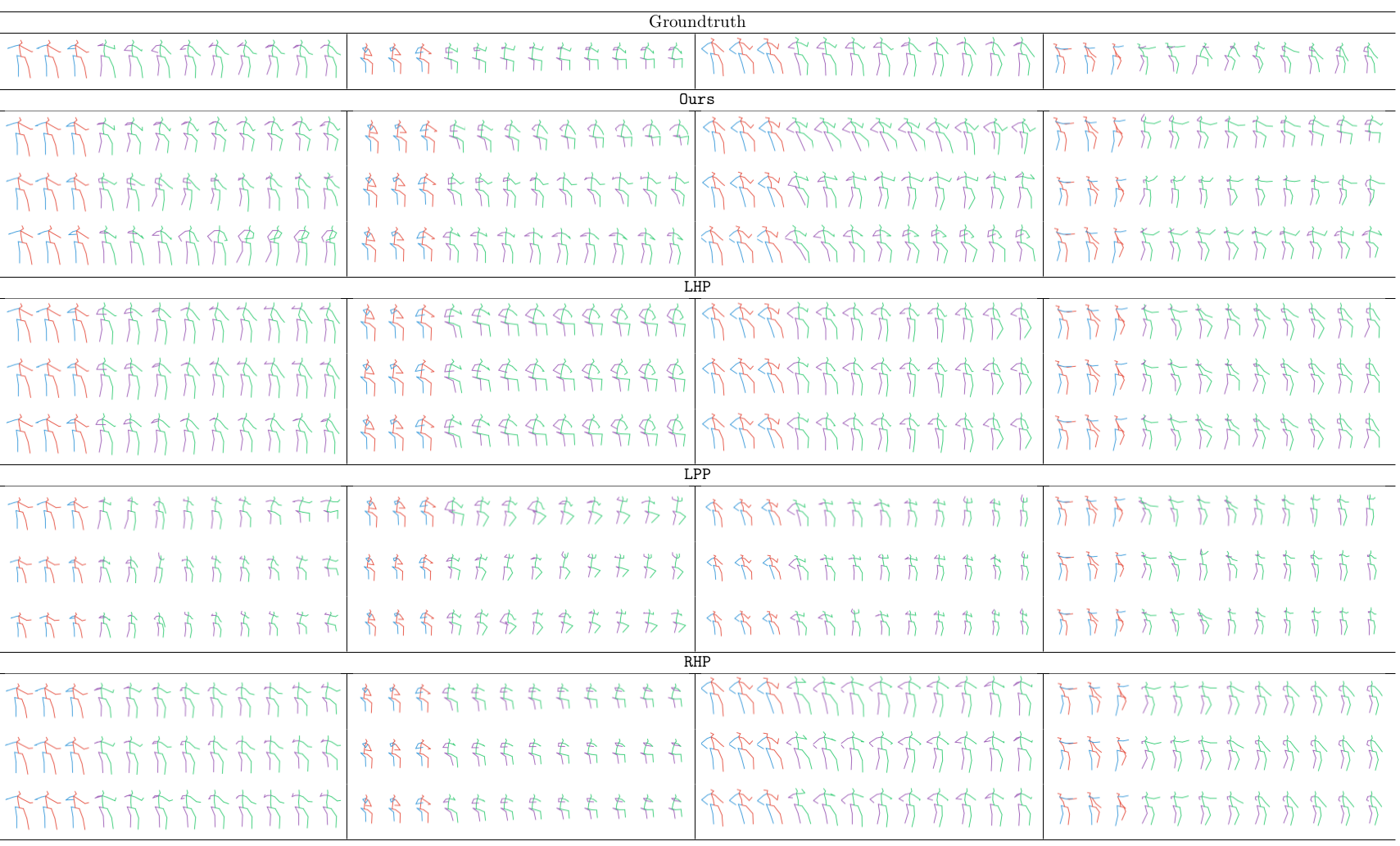}
    \caption{Qualitative visualization of diverse motions generated by our model and by the baselines. Each block of columns shows the results for one observation (the first three poses of each sequence). The first row corresponds to the ground-truth motion and the other rows illustrate  multiple motions generated by each method (better seen when zoomed in).}
    \label{fig:qaulitative}
\end{figure*}

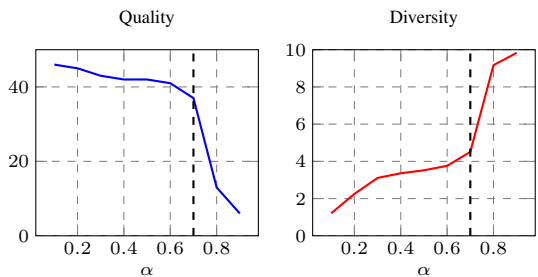
\begin{figure}
    \centering
    \begin{tabular}{@{ }c@{ }c}
        \pgfplotsset{width=0.26\textwidth,height=0.18\textheight, compat=1.3, legend columns=1,  legend style={at={(-0.5,1)},anchor=north}, grid style={dashed,gray}}  

\begin{tikzpicture}
\begin{axis}[
%   axis y line*=left,
  ymin=0, ymax=50,
   xlabel=$\alpha$,
  grid=both,
   title={Quality}
%   style={thick}
]
\scriptsize
\addplot[blue, style={thick}]
  coordinates{
(	0.1	,    46 )
(	0.2	,	45	)
(	0.3	,	43	)
(	0.4	,	42	)
(	0.5	,	42	)
(	0.6	,	41	)
(	0.7	,	37	)
(	0.8	,	13	)
(	0.9	,	6	)
}; 

\addplot +[mark=none, black, dashed, style={thick}] coordinates {(0.7, 0) (0.7, 50)};
\end{axis}

\end{tikzpicture} & \pgfplotsset{width=0.26\textwidth,height=0.18\textheight, compat=1.3, legend columns=1,  legend style={at={(-0.5,1)},anchor=north}, grid style={dashed,gray}}

\begin{tikzpicture}

\begin{axis}[
%   axis y line*=left,
  ymin=0, ymax=10,
   xlabel=$\alpha$,
  grid=both,
   title={Diversity}
%   style={thick}
]
\scriptsize
\addplot[red, style={thick}]
  coordinates{
(	0.1	,    1.21   )
(	0.2	,	2.25	)
(	0.3	,	3.11	)
(	0.4	,	3.36	)
(	0.5	,	3.52	)
(	0.6	,	3.76	)
(	0.7	,	4.51	)
(	0.8	,	9.17	)
(	0.9	,	9.83	)
}; 
\addplot +[mark=none, black, dashed, style={thick}] coordinates {(0.7, 0) (0.7, 10)};

\end{axis}
\end{tikzpicture}
    \end{tabular}
    
    \caption{Quality and the diversity of the motions generated with our approach as a function of $\alpha$.
    Note that with $\alpha>0.7$, diversity increases significantly, but this diversity is the result of poor-quality motions.}
    \label{fig:tradeoff}
\end{figure}

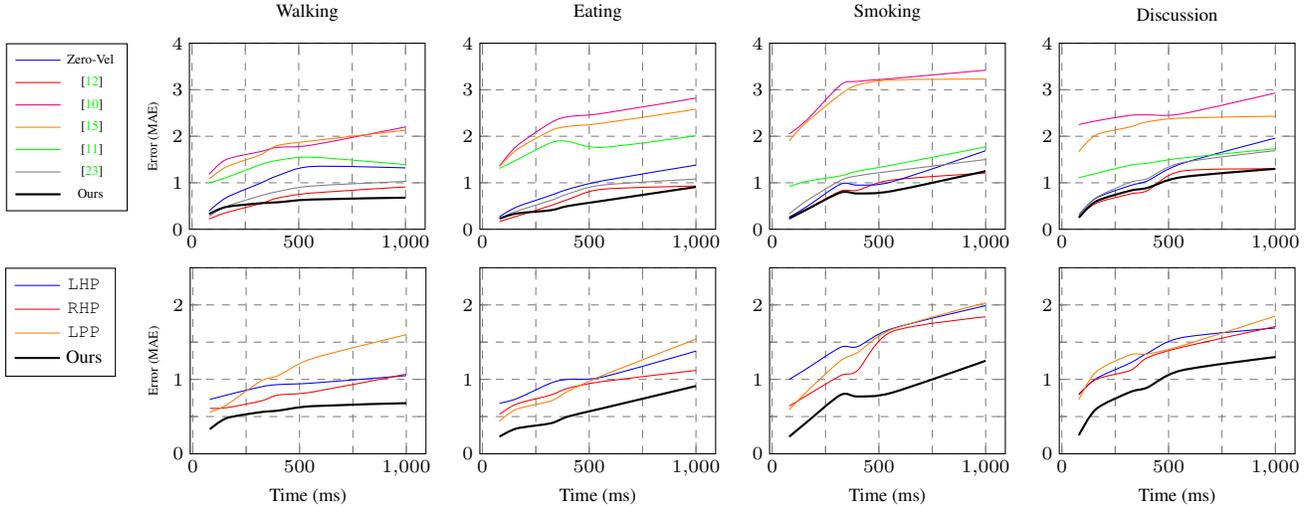
\begin{figure*}[t]
    \centering
    \begin{tabular}{@{ }c@{ }c@{ }c@{ }c}
         \pgfplotsset{width=0.27\textwidth,height=0.18\textheight, compat=1.3, legend columns=1,  legend style={at={(-0.52,1)},anchor=north}, grid style={dashed,gray}}  

\begin{tikzpicture}
\tiny
\begin{axis}[
%   axis y line*=left,
  ymin=0, ymax=4,
%   xlabel=Time (ms),
  ylabel={\tiny Error (MAE)},
  grid=both,
   minor x tick num=1,
  title={Walking},
  %style={thick}
]
\scriptsize

\addplot[smooth,blue]
  coordinates{
(	80	,	0.39	)
(	160	,	0.68	)
(	320	,	0.99	)
(	400	,	1.15	)
(	560	,	1.35	)
(	1000	,	1.32	)

};
\label{Zero-Velocity}
\addlegendentry{\tiny Zero-Vel}

\addplot[smooth,red]
  coordinates{
(	80	,	0.22	)
(	160	,	0.36	)
(	320	,	0.55	)
(	400	,	0.67	)
(	560	,	0.78	)
(	1000	,	0.91	)

}; \label{AGED}
\addlegendentry{\tiny ~\cite{gui2018adversarial}}

% \addplot[smooth,cyan]
%   coordinates{
% (	80	,	1.3	)
% (	160	,	1.56	)
% (	320	,	1.84	)
% (	400	,	1.93	)
% (	560	,	2	)
% (	1000	,	2.38	)

% }; \label{ERD}
% \addlegendentry{\tiny ~\cite{fragkiadaki2015recurrent}}

\addplot[smooth,magenta]
  coordinates{
(	80	,	1.18	)
(	160	,	1.5	)
(	320	,	1.67	)
(	400	,	1.76	)
(	560	,	1.81	)
(	1000	,	2.2	)

}; \label{LSTM-3LR}
\addlegendentry{\tiny ~\cite{fragkiadaki2015recurrent}}

\addplot[smooth,orange]
  coordinates{
(	80	,	1.08	)
(	160	,	1.34	)
(	320	,	1.6	)
(	400	,	1.8	)
(	560	,	1.9	)
(	1000	,	2.13	)

}; \label{SRNN}
\addlegendentry{\tiny ~\cite{jain2016structural}}

\addplot[smooth,green]
  coordinates{
(	80	,	1	)
(	160	,	1.11	)
(	320	,	1.39	)
(	400	,	1.48	)
(	560	,	1.55	)
(	1000	,	1.39	)

}; \label{DAE-LSTM}
\addlegendentry{\tiny ~\cite{ghosh2017learning}}

\addplot[smooth,gray]
  coordinates{
(	80	,	0.28	)
(	160	,	0.49	)
(	320	,	0.72	)
(	400	,	0.81	)
(	560	,	0.93	)
(	1000	,	1.03	)

}; \label{GRU}
\addlegendentry{\tiny ~\cite{martinez2017human}}

% \addplot[smooth,black, style={thin}]
%   coordinates{
% (	80	,	0.31	)
% (	160	,	0.39	)
% (	320	,	0.57	)
% (	400	,	0.64	)
% (	560	,	0.74	)
% (	1000	,	0.83	)

% }; \label{Ours}
% \addlegendentry{\tiny Ours k=100}

\addplot[smooth,black, style={thick}]
  coordinates{
(	80	,	0.33	)
(	160	,	0.48	)
(	320	,	0.56	)
(	400	,	0.58	)
(	560	,	0.64	)
(	1000	,	0.68	)

}; \label{Ours}
\addlegendentry{\tiny Ours}

% \addplot[smooth,black, dashed, style={thick}]
%   coordinates{
% (	80	,	0.35	)
% (	160	,	0.51	)
% (	320	,	0.60	)
% (	400	,	0.63	)
% (	560	,	0.74	)
% (	1000	,	0.78	)

% }; \label{Ours}
% \addlegendentry{\tiny Ours new}

\end{axis}

\end{tikzpicture} &
         \pgfplotsset{width=0.27\textwidth,height=0.18\textheight, compat=1.3, legend columns=1,  legend style={at={(-0.5,1)},anchor=north}, grid style={dashed,gray}}

\begin{tikzpicture}
\tiny
\begin{axis}[
%   axis y line*=left,
  ymin=0, ymax=4,
%   xlabel=Time (ms),
%   ylabel={\scriptsize Error (MAE)},
  grid=both,
 minor x tick num=1,
  title={Eating}
  %style={thick}
]
\scriptsize
\addplot[smooth,red]
  coordinates{
(	80	,	0.17	)
(	160	,	0.28	)
(	320	,	0.51	)
(	400	,	0.64	)
(	560	,	0.86	)
(	1000	,	0.93	)

}; \label{AGED}
%\addlegendentry{AGED}

\addplot[smooth,blue]
  coordinates{
(	80	,	0.27	)
(	160	,	0.48	)
(	320	,	0.73	)
(	400	,	0.86	)
(	560	,	1.04	)
(	1000	,	1.38	)

}; \label{Zero-Velocity}
%\addlegendentry{Zero-Velocity}

% \addplot[smooth,cyan]
%   coordinates{
% (	80	,	1.66	)
% (	160	,	1.93	)
% (	320	,	2.28	)
% (	400	,	2.32	)
% (	560	,	2.36	)
% (	1000	,	2.41	)

% }; \label{ERD}
% %\addlegendentry{ERD}

\addplot[smooth,magenta]
  coordinates{
(	80	,	1.36	)
(	160	,	1.79	)
(	320	,	2.29	)
(	400	,	2.42	)
(	560	,	2.49	)
(	1000	,	2.82	)

}; \label{LSTM-3LR}
%\addlegendentry{LSTM-3LR}

\addplot[smooth,orange]
  coordinates{
(	80	,	1.35	)
(	160	,	1.71	)
(	320	,	2.12	)
(	400	,	2.21	)
(	560	,	2.28	)
(	1000	,	2.58	)

}; \label{SRNN}
%\addlegendentry{SRNN}

\addplot[smooth,green]
  coordinates{
(	80	,	1.31	)
(	160	,	1.49	)
(	320	,	1.86	)
(	400	,	1.89	)
(	560	,	1.76	)
(	1000	,	2.01	)

}; \label{DAE-LSTM}
%\addlegendentry{DAE-LSTM}

\addplot[smooth,gray]
  coordinates{
(	80	,	0.23	)
(	160	,	0.39	)
(	320	,	0.62	)
(	400	,	0.76	)
(	560	,	0.95	)
(	1000	,	1.08	)

}; \label{GRU}
%\addlegendentry{GRU}

% \addplot[smooth,black, style={thin}]
%   coordinates{
% (	80	,	0.24	)
% (	160	,	0.33	)
% (	320	,	0.52	)
% (	400	,	0.64	)
% (	560	,	0.70	)
% (	1000	,	1.0	)

% }; \label{Ours}
% %\addlegendentry{Ours}
\addplot[smooth,black, style={thick}]
  coordinates{
(	80	,	0.23	)
(	160	,	0.34	)
(	320	,	0.41	)
(	400	,	0.5	)
(	560	,	0.61	)
(	1000	,	0.91	)

}; \label{Ours}

% \addplot[smooth,dashed, black, style={thick}]
%   coordinates{
% (	80		0.1	)
% (	160	,	0.1	)
% (	320	,	0.1	)
% (	400	,	0.1	)
% (	560	,	0.1	)
% (	1000	,	0.1	)

% }; 

% \addplot[smooth,dotted, black, style={thick}]
%   coordinates{
% (	80		0.2	)
% (	160	,	0.2	)
% (	320	,	0.2	)
% (	400	,	0.2	)
% (	560	,	0.2)
% (	1000	,	0.2	)

% }; \label{Ours}
% \addplot[smooth,black, dashed, style={thick}]
%   coordinates{
% (	80	,	0.29	)
% (	160	,	0.38	)
% (	320	,	0.47	)
% (	400	,	0.59	)
% (	560	,	0.68	)
% (	1000	,	1.04	)

% }; \label{Ours new}

\end{axis}

\end{tikzpicture} &
         \pgfplotsset{width=0.27\textwidth,height=0.18\textheight, compat=1.3, legend columns=1,  legend style={at={(-0.5,1)},anchor=north}, grid style={dashed,gray}}

\begin{tikzpicture}
\tiny
\begin{axis}[
%   axis y line*=left,
  ymin=0, ymax=4,
%   xlabel=Time (ms),
%   ylabel={\scriptsize Error (MAE)},
  grid=both,
   minor x tick num=1,
  title={Smoking}
  %style={thick}
]
\scriptsize
\addplot[smooth,red]
  coordinates{
(	80	,	0.27	)
(	160	,	0.43	)
(	320	,	0.82	)
(	400	,	0.84	)
(	560	,	1.06	)
(	1000	,	1.21	)

}; \label{AGED}
%\addlegendentry{AGED}

\addplot[smooth,blue]
  coordinates{
(	80	,	0.26	)
(	160	,	0.48	)
(	320	,	0.97	)
(	400	,	0.95	)
(	560	,	1.02	)
(	1000	,	1.69	)

}; \label{Zero-Velocity}
%\addlegendentry{Zero-Velocity}

% \addplot[smooth,cyan]
%   coordinates{
% (	80	,	2.34	)
% (	160	,	2.74	)
% (	320	,	3.73	)
% (	400	,	3.70	)
% (	560	,	3.68	)
% (	1000	,	3.82	)

% }; \label{ERD}
% %\addlegendentry{ERD}

\addplot[smooth,magenta]
  coordinates{
(	80	,	2.05	)
(	160	,	2.34	)
(	320	,	3.10	)
(	400	,	3.18	)
(	560	,	3.24	)
(	1000	,	3.42	)

}; \label{LSTM-3LR}
%\addlegendentry{LSTM-3LR}

\addplot[smooth,orange]
  coordinates{
(	80	,	1.90	)
(	160	,	2.30	)
(	320	,	2.90	)
(	400	,	3.10	)
(	560	,	3.21	)
(	1000	,	3.23	)

}; \label{SRNN}
%\addlegendentry{SRNN}

\addplot[smooth,green]
  coordinates{
(	80	,	0.92	)
(	160	,	1.03	)
(	320	,	1.15	)
(	400	,	1.25	)
(	560	,	1.38	)
(	1000	,	1.77	)

}; \label{DAE-LSTM}
%\addlegendentry{DAE-LSTM}

\addplot[smooth,gray]
  coordinates{
(	80	,	0.33	)
(	160	,	0.61	)
(	320	,	1.05	)
(	400	,	1.15	)
(	560	,	1.25	)
(	1000	,	1.5	)

}; \label{GRU}
%\addlegendentry{GRU}

% \addplot[smooth,black, style={thin}]
%   coordinates{
% (	80	,	0.26	)
% (	160	,	0.42	)
% (	320	,	0.85	)
% (	400	,	0.88	)
% (	560	,	0.96	)
% (	1000	,	1.25	)

% }; \label{Ours}

\addplot[smooth,black, style={thick}]
  coordinates{
(	80	,	0.23	)
(	160	,	0.42	)
(	320	,	0.79	)
(	400	,	0.77	)
(	560	,	0.82	)
(	1000	,	1.25	)

}; \label{Ours}

% \addplot[smooth,black, dashed, style={thick}]
%   coordinates{
% (	80	,	0.33	)
% (	160	,	0.52	)
% (	320	,	0.87	)
% (	400	,	0.88	)
% (	560	,	0.92	)
% (	1000	,	1.41	)

% }; \label{Ours new}
%\addlegendentry{Ours}

% \addplot[smooth,dashed, black, style={thick}]
%   coordinates{
% (	80		0.1	)
% (	160	,	0.1	)
% (	320	,	0.1	)
% (	400	,	0.1	)
% (	560	,	0.1	)
% (	1000	,	0.1	)

% }; 

% \addplot[smooth,dotted, black, style={thick}]
%   coordinates{
% (	80		0.2	)
% (	160	,	0.2	)
% (	320	,	0.2	)
% (	400	,	0.2	)
% (	560	,	0.2)
% (	1000	,	0.2	)

% }; \label{Ours}

\end{axis}

\end{tikzpicture} &
         \pgfplotsset{width=0.27\textwidth,height=0.18\textheight, compat=1.3, legend columns=1,  legend style={at={(-0.5,1)},anchor=north}, grid style={dashed,gray}}

\begin{tikzpicture}
\tiny
\begin{axis}[
%   axis y line*=left,
  ymin=0, ymax=4,
%   xlabel=Time (ms),
%   ylabel={\scriptsize Error (MAE)},
  grid=both,
  minor x tick num=1,
  title={Discussion}
  %style={thick}
]
\scriptsize
\addplot[smooth,red]
  coordinates{
(	80	,	0.27	)
(	160	,	0.56	)
(	320	,	0.76	)
(	400	,	0.83	)
(	560	,	1.25	)
(	1000	,	1.30	)

}; \label{AGED}
%\addlegendentry{AGED}

\addplot[smooth,blue]
  coordinates{
(	80	,	0.31	)
(	160	,	0.67	)
(	320	,	0.94	)
(	400	,	1.04	)
(	560	,	1.41	)
(	1000	,	1.96	)

}; \label{Zero-Velocity}
%\addlegendentry{Zero-Velocity}

% \addplot[smooth,cyan]
%   coordinates{
% (	80	,	2.67	)
% (	160	,	2.97	)
% (	320	,	3.23	)
% (	400	,	3.32	)
% (	560	,	3.47	)
% (	1000	,	2.92	)

% }; \label{ERD}
% %\addlegendentry{ERD}

\addplot[smooth,magenta]
  coordinates{
(	80	,	2.25	)
(	160	,	2.33	)
(	320	,	2.45	)
(	400	,	2.46	)
(	560	,	2.48	)
(	1000	,	2.93	)

}; \label{LSTM-3LR}
%\addlegendentry{LSTM-3LR}

\addplot[smooth,orange]
  coordinates{
(	80	,	1.67	)
(	160	,	2.03	)
(	320	,	2.20	)
(	400	,	2.31	)
(	560	,	2.39	)
(	1000	,	2.43	)

}; \label{SRNN}
%\addlegendentry{SRNN}

\addplot[smooth,green]
  coordinates{
(	80	,	1.11	)
(	160	,	1.20	)
(	320	,	1.38	)
(	400	,	1.42	)
(	560	,	1.53	)
(	1000	,	1.73	)

}; \label{DAE-LSTM}
%\addlegendentry{DAE-LSTM}

\addplot[smooth,gray]
  coordinates{
(	80	,	0.31	)
(	160	,	0.68	)
(	320	,	1.01	)
(	400	,	1.09	)
(	560	,	1.43	)
(	1000	,	1.69	)

}; \label{GRU}
%\addlegendentry{GRU}

% \addplot[smooth,black, style={thin}]
%   coordinates{
% (	80,		0.29	)
% (	160	,	0.60	)
% (	320	,	0.82	)
% (	400	,	0.89	)
% (	560	,	1.13	)
% (	1000	,	1.44	)

% }; \label{Ours}

\addplot[smooth,black, style={thick}]
  coordinates{
(	80,		0.25	)
(	160	,	0.60	)
(	320	,	0.83	)
(	400	,	0.89	)
(	560	,	1.12	)
(	1000	,	1.30	)

}; \label{Ours}

% \addplot[smooth,black,dashed, style={thick}]
%   coordinates{
% (	80,		0.32	)
% (	160	,	0.63	)
% (	320	,	0.83	)
% (	400	,	0.89	)
% (	560	,	1.17	)
% (	1000	,	1.44	)

% }; \label{Ours new}
%\addlegendentry{Ours}

% \addplot[smooth,dashed, black, style={thick}]
%   coordinates{
% (	80		0.1	)
% (	160	,	0.1	)
% (	320	,	0.1	)
% (	400	,	0.1	)
% (	560	,	0.1	)
% (	1000	,	0.1	)

% }; 

% \addplot[smooth,dotted, black, style={thick}]
%   coordinates{
% (	80		0.2	)
% (	160	,	0.2	)
% (	320	,	0.2	)
% (	400	,	0.2	)
% (	560	,	0.2)
% (	1000	,	0.2	)

% }; \label{Ours}

\end{axis}

\end{tikzpicture} \\
         
         \pgfplotsset{width=0.27\textwidth,height=0.18\textheight, compat=1.3, legend columns=1,  legend style={at={(-0.55,1)},anchor=north}, grid style={dashed,gray}}  

\begin{tikzpicture}
\tiny
\begin{axis}[
%   axis y line*=left,
  ymin=0, ymax=2.5,
   xlabel=Time (ms),
  ylabel={\tiny Error (MAE)},
  grid=both,
  minor tick num=1
%   title={Walking}
  %style={thick}
]
\scriptsize

\addplot[smooth,blue]
  coordinates{
(	80	,	0.73	)
(	160	,	0.79	)
(	320	,	0.90	)
(	400	,	0.93	)
(	560	,	0.95	)
(	1000	,	1.05	)

}; \label{LHP}
\addlegendentry{\scriptsize \texttt{LHP}}

\addplot[smooth,red]
  coordinates{
(	80	,	0.61	)
(	160	,	0.62	)
(	320	,	0.71	)
(	400	,	0.79	)
(	560	,	0.83	)
(	1000	,	1.07	)

}; \label{RHP}
\addlegendentry{\scriptsize \texttt{RHP}}

\addplot[smooth,orange]
  coordinates{
(	80	,	0.56	)
(	160	,	0.66	)
(	320	,	0.98	)
(	400	,	1.05	)
(	560	,	1.28	)
(	1000	, 1.60		)

}; \label{ERD}
\addlegendentry{\scriptsize \texttt{LPP}}

% \addplot[smooth,black, style={thick}]
%   coordinates{
% (	80	,	0.31	)
% (	160	,	0.39	)
% (	320	,	0.57	)
% (	400	,	0.64	)
% (	560	,	0.74	)
% (	1000	,	0.83	)

% }; \label{Ours}
% \addlegendentry{\scriptsize Ours}
\addplot[smooth,black, style={thick}]
  coordinates{
(	80	,	0.33	)
(	160	,	0.48	)
(	320	,	0.56	)
(	400	,	0.58	)
(	560	,	0.64	)
(	1000	,	0.68	)

}; \label{Ours}
\addlegendentry{\scriptsize Ours}

\end{axis}

\end{tikzpicture} &
         \pgfplotsset{width=0.27\textwidth,height=0.18\textheight, compat=1.3, legend columns=1,  legend style={at={(-0.52,1)},anchor=north}, grid style={dashed,gray}}  

\begin{tikzpicture}
\tiny
\begin{axis}[
%   axis y line*=left,
  ymin=0, ymax=2.5,
   xlabel=Time (ms),
  grid=both,
  minor tick num=1
%   title={Eating}
  %style={thick}
]
\scriptsize

\addplot[smooth,blue]
  coordinates{
(	80	,	0.68	)
(	160	,	0.74	)
(	320	,	0.95	)
(	400	,	1.00	)
(	560	,	1.03	)
(	1000	,	1.38	)

}; \label{LHP}
% \addlegendentry{\scriptsize \texttt{LHP}}

\addplot[smooth,red]
  coordinates{
(	80	,	0.53	)
(	160	,	0.67	)
(	320	,	0.79	)
(	400	,	0.88	)
(	560	,	0.97	)
(	1000	,	1.12	)

}; \label{RHP}
% \addlegendentry{\scriptsize \texttt{RHP}}

\addplot[smooth,orange]
  coordinates{
(	80	,	0.44	)
(	160	,	0.60	)
(	320	,	0.71	)
(	400	,	0.84	)
(	560	,	1.05	)
(	1000	, 1.54		)

}; \label{ERD}
% \addlegendentry{\scriptsize \texttt{LPP}}

% \addplot[smooth,black, style={thick}]
%   coordinates{
% (	80	,	0.24	)
% (	160	,	0.33	)
% (	320	,	0.52	)
% (	400	,	0.64	)
% (	560	,	0.70	)
% (	1000	,	1.0	)

% }; \label{Ours}

\addplot[smooth,black, style={thick}]
  coordinates{
(	80	,	0.23	)
(	160	,	0.34	)
(	320	,	0.41	)
(	400	,	0.5	)
(	560	,	0.61	)
(	1000	,	0.91	)

}; \label{Ours}

\end{axis}

\end{tikzpicture} &
         \pgfplotsset{width=0.27\textwidth,height=0.18\textheight, compat=1.3, legend columns=1,  legend style={at={(-0.52,1)},anchor=north}, grid style={dashed,gray}}  

\begin{tikzpicture}
\tiny
\begin{axis}[
%   axis y line*=left,
  ymin=0, ymax=2.5,
   xlabel=Time (ms),
  grid=both,
  minor tick num=1
%   title={Eating}
  %style={thick}
]
\scriptsize

\addplot[smooth,blue]
  coordinates{
(	80	,	1.0	)
(	160	,	1.14	)
(	320	,	1.43	)
(	400	,	1.44	)
(	560	,	1.68	)
(	1000	,	1.99	)

}; \label{LHP}
% \addlegendentry{\scriptsize \texttt{LHP}}

\addplot[smooth,red]
  coordinates{
(	80	,	0.64	)
(	160	,	0.78	)
(	320	,	1.05	)
(	400	,	1.12	)
(	560	,	1.64	)
(	1000	,	1.84	)

}; \label{RHP}
% \addlegendentry{\scriptsize \texttt{RHP}}

\addplot[smooth,orange]
  coordinates{
(	80	,	0.59	)
(	160	,	0.83	)
(	320	,	1.25	)
(	400	,	1.36	)
(	560	,	1.67	)
(	1000	, 2.03		)

}; \label{ERD}
% \addlegendentry{\scriptsize \texttt{LPP}}

% \addplot[smooth,black, style={thick}]
%   coordinates{
% (	80	,	0.26	)
% (	160	,	0.42	)
% (	320	,	0.85	)
% (	400	,	0.88	)
% (	560	,	0.96	)
% (	1000	,	1.25	)

% }; \label{Ours}

\addplot[smooth,black, style={thick}]
  coordinates{
(	80	,	0.23	)
(	160	,	0.42	)
(	320	,	0.79	)
(	400	,	0.77	)
(	560	,	0.82	)
(	1000	,	1.25	)

}; \label{Ours}

\end{axis}

\end{tikzpicture} &
         \pgfplotsset{width=0.27\textwidth,height=0.18\textheight, compat=1.3, legend columns=1,  legend style={at={(-0.52,1)},anchor=north}, grid style={dashed,gray}}  

\begin{tikzpicture}
\tiny
\begin{axis}[
%   axis y line*=left,
  ymin=0, ymax=2.5,
   xlabel=Time (ms),
  grid=both,
  minor tick num=1
%   title={Eating}
  %style={thick}
]
\scriptsize

\addplot[smooth,blue]
  coordinates{
(	80	,	0.80	)
(	160	,	1.01	)
(	320	,	1.22	)
(	400	,	1.35	)
(	560	,	1.56	)
(	1000	,	1.69	)

}; \label{LHP}
% \addlegendentry{\scriptsize \texttt{LHP}}

\addplot[smooth,red]
  coordinates{
(	80	,	0.79	)
(	160	,	1.00	)
(	320	,	1.12	)
(	400	,	1.29	)
(	560	,	1.43	)
(	1000	,	1.71	)

}; \label{RHP}
% \addlegendentry{\scriptsize \texttt{RHP}}

\addplot[smooth,orange]
  coordinates{
(	80	,	0.73	)
(	160	,	1.10	)
(	320	,	1.33	)
(	400	,	1.34	)
(	560	,	1.45	)
(	1000	, 1.85		)

}; \label{LPP}
% \addlegendentry{\scriptsize \texttt{LPP}}

% \addplot[smooth,black, style={thick}]
%   coordinates{
% (	80,		0.29	)
% (	160	,	0.60	)
% (	320	,	0.82	)
% (	400	,	0.89	)
% (	560	,	1.13	)
% (	1000	,	1.44	)

% }; \label{Ours}

% \addplot[smooth,black, style={thick}]
%   coordinates{
% (	80,		0.29	)
% (	160	,	0.60	)
% (	320	,	0.82	)
% (	400	,	0.89	)
% (	560	,	1.13	)
% (	1000	,	1.44	)

% }; \label{Ours}
\addplot[smooth,black, style={thick}]
  coordinates{
(	80,		0.25	)
(	160	,	0.60	)
(	320	,	0.83	)
(	400	,	0.89	)
(	560	,	1.12	)
(	1000	,	1.30	)

}; \label{Ours}

\end{axis}

\end{tikzpicture} \\
         
    \end{tabular}
    \caption{Mean angle error (MAE) for the Human 3.6M actions commonly used to report long-term prediction results. \textbf{(Top)} We compare the best of $K$ motions generated by our approach with the state-of-the-art deterministic baselines. Note that, while our approach does exploits knowledge of the action during neither training nor inference (unlike some of the baselines), it performs on par with the state-of-the-art deterministic baselines. \textbf{(Bottom)} We compare our approach with the state-of-the-art stochastic baselines. Note that the results for each stochastic baseline were obtained from the best of $K$ generated motions.
    }
    \label{fig:long_term}
\end{figure*}

\begin{figure*}[!t]
    \centering
    \includegraphics[width=0.95\textwidth]{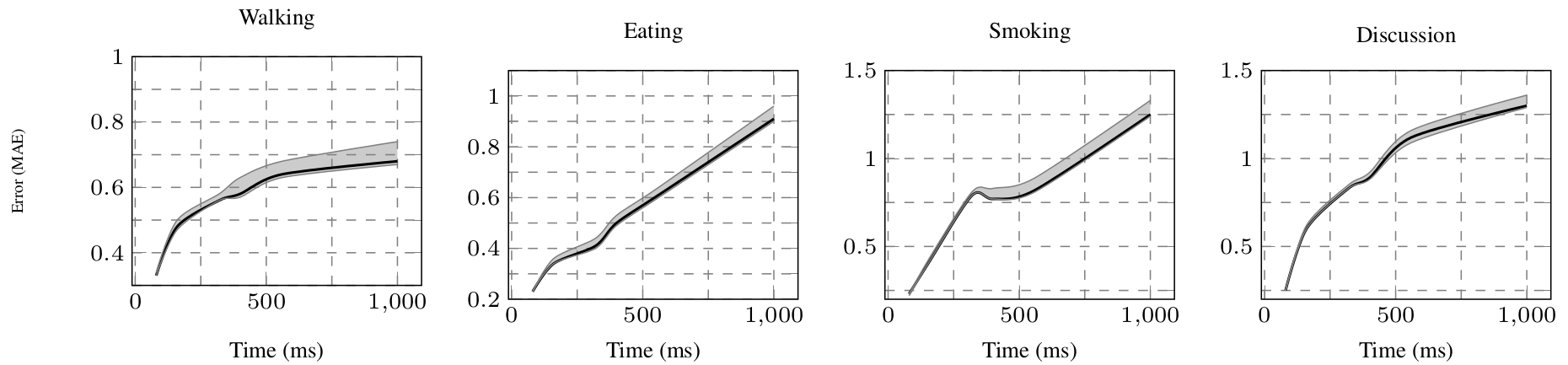}
    \caption{Effect of $K$ in the MAE of our actions of the Human3.6M dataset. Note, the bold black one is the best of $K=50$ motions, and the shaded area indicates the region between best of $K=1$ and $K=500$.}
    \label{fig:k_effect}
\end{figure*}

\paragraph{\texttt{RHP}.} As for \texttt{LHP}, Fig.~\ref{fig:QD_plot} evidences the limited diversity of the motions produced by \texttt{RHP} despite its use of random noise during inference. Note that the authors of~\cite{barsoum2018hp}  mentioned in their paper that the random noise was added to the hidden state. Only by studying their publicly available code\footnote{\texttt{https://github.com/ebarsoum/hpgan}} did we understand the precise way this combination was done. In fact, the addition relies on a parametric, linear transformation of the noise vector. That is, the perturbed hidden state is obtained as
\begin{align}
    h_{perturbed} = h_{original} + W^{z\rightarrow h} z\;.
\end{align}
Because the parameters $W^{z\rightarrow h}$ are \emph{learnt}, the model has the flexibility to ignore $z$, which causes the behavior observed in Figs.~\ref{fig:QD_plot} and~\ref{fig:baseline_RHP}. Note that the authors of~\cite{barsoum2018hp} acknowledged that, despite their best efforts, they noticed very little variations between predictions obtained with different $z$ values. Since the perturbation is ignored, however, the quality of the generated motions is high. By depicting \texttt{RHP} in a region of high quality but low diversity, the human evaluation results in Fig.~\ref{fig:human_eval} again match those of our classifier-based quality metric.

\paragraph{\texttt{LPP}.} As can be seen in Fig.~\ref{fig:QD_plot}, \texttt{LPP} produces motions with higher diversity than \texttt{LHP} and \texttt{RHP}, but of much lower quality. The main reason behind this is that the random vectors that are concatenated to the poses at each time-step are sampled independently of each other, which translates to discontinuities in the generated motions. This problem might be mitigated by sampling the noise in a time-dependent, autoregressive manner, as in~\cite{videoFlow} for video generation. Doing so, however, goes beyond the scope of our analysis. When it comes to human evaluation, Fig.\ref{fig:human_eval} further confirms that \texttt{LPP}'s results lie in a low-quality, medium-diversity region.

\paragraph{\texttt{Ours}.} The goal of our mix-and-match perturbations was to make it hard for the model to decouple the random vector from the deterministic hidden state information. The success of our approach is confirmed by Fig.~\ref{fig:QD_plot}. Our model generates diverse motions, even after a long training time, and the quality of these motions is high. While this quality is slightly lower than that of \texttt{LHP} and \texttt{RHP} when looking at our classifier-based metric, it is rated higher by humans, as can be verified from Fig.~\ref{fig:human_eval}. We believe that this discrepancy is related to the binary classifier memorizing the ground-truth motions and thus not generalizing to the large diversity of motions generated by our model. As such, human evaluation still nicely complements our less expensive automatic one. Altogether, these results confirm the ability of our approach to generate highly diverse yet realistic motions. In Fig.~\ref{fig:qaulitative}, we further evidence this qualitatively by providing samples obtained by our approach for four different input sequences, as well as  samples from the baselines. Note that, for each input sequence, we produce large, yet natural variations of future poses.

Note that our approach depends on the parameter $\alpha$, which defines the amount of randomness used in our mix-and-match perturbations. In Fig.~\ref{fig:tradeoff}, we report the quality and diversity of our results when varying $\alpha\in\{0.1, \dots, 0.9\}$. Note that these plots show a trade-off between quality and diversity. This is to be expected since, by aiming to increase diversity, the resulting motion will become unrealistic. Nevertheless, our results can be seen to be highly diverse and of high quality for a wide range of values, i.e., by setting $\alpha\in[0.3,0.7]$. Note that, $\alpha$ is the only model-related hyper-parameter of the Mix-and-Match. The quality and diversity metrics are monotonic functions of $\alpha$, thus, one can choose a proper $\alpha$ given a task. Note that, using $\alpha=0.2$, our method still achieves a SoTA diversity of 2.25 with a higher quality of 45.0\%, however, for the sake of fair comparison, we use the default value of $\alpha=0.5$.

\subsection{Comparison with the State of the Art}
\label{sec:sota}
We now compare the results of our approach with those obtained using the state-of-the-art deterministic motion prediction methods~\cite{martinez2017human,jain2016structural,gui2018few,pavllo2018quaternet,pavllo2019modeling,fragkiadaki2015recurrent,gui2018adversarial} for long-term motion prediction, i.e., up to 1000ms. For this experiment, following previous work~\cite{fragkiadaki2015recurrent,pavllo2019modeling,pavllo2018quaternet,martinez2017human,gui2018adversarial}, we model velocity instead of pose, and do the same for the stochastic baselines. This is achieved by adding a residual connection to the motion decoder.
We then report the standard metric, i.e., the Euclidean distance between the predicted and ground-truth Euler angles. To evaluate this metric for our method which generates multiple, diverse predictions, we make use of the best sample among the $K$ generated ones with $K=50$ for the stochastic baselines and for our approach (i.e., the S-MSE metric~\cite{yan2018mt}).
In other words, we aim to show that, among the $K$ motions we generate, at least one is close to the ground truth. As can be seen in the top portion of Fig.~\ref{fig:long_term}, our approach yields errors comparable to the best-performing baselines, despite their use of more complex architectures and strong losses, such as the adversarial loss used in~\cite{gui2018adversarial}. Note that, unlike some of the baselines~\cite{gui2018adversarial,jain2016structural,martinez2017human}, our model requires knowing the action class during neither training nor inference.

In the bottom portion of Fig.~\ref{fig:long_term}, we further compare our best estimate with that of the other stochastic baselines, \texttt{LPH}, \texttt{RHP}, and \texttt{LPP}, using the best of $K$ motions in all cases. Note that, by providing better diversity, our approach outperforms these baselines. 

\section{Evaluating the Effect of $K$}
In the main paper, we used $K=50$ to compare our approach with the state-of-the-art deterministic and stochastic baselines. Here, we provide an ablation study on the effect of $K$. To this end, we provide results when using $K=1$ to $K=500$. In Fig.~\ref{fig:k_effect}, we plot the results with $K=50$ as bold black lines, and the shaded area covers the results obtained with $K=1$ to $K=500$. While smaller values of $K$ yield large errors, the difference between $K=50$ and $K=500$ is very small (barely visible in most cases).

 \section{Conclusion}
In this paper, we have proposed an effective way of perturbing the hidden state of an RNN such that it becomes capable of learning the multiple modes of human motions. Our evaluation of quality and diversity, based on both new quantitative metrics and human judgment, have evidenced that our approach outperforms the state-of-the-art stochastic methods. 
Generating diverse plausible motions given a short sequence of observations has many applications, especially when the motions are generated in an action-agnostic manner. For instance, our model can be used for human action forecasting~\cite{aliakbarian2016deep,rodriguez2018action,shi2018action,Aliakbarian_2017_ICCV,aliakbarian2018viena}, where one seeks to anticipate the action as early as possible. 
It can also be employed for motion inpainting, where, given partial observations, one aims to generate multiple in-between solutions.
In the future, we will therefore investigate the use of our approach in such applications.

{\small
\bibliographystyle{ieee}
\bibliography{egpaper_for_review}
}

\end{document}

%%%%%%%%% TITLE
\title{Learning Variations in Human Motion via Mix-and-Match Perturbation\\Supplementary Material}

\author{Mohammad Sadegh Aliakbarian$^{1,2,3}$\thanks{Work partially done during internship with Qualcomm AI Research. Contact: \texttt{sadegh.aliakbarian@anu.edu.au}} , Fatemeh Sadat Saleh$^{1,2}$, Mathieu Salzmann$^{4}$,\\Lars Petersson$^{1,3}$,Stephen Gould$^{1,2}$, Amirhossein Habibian$^{5}$\\
$^{1}$ANU, $^{2}$ACRV, $^{3}$Data61-CSIRO, $^{4}$CVLab-EPFL, $^{5}$Qualcomm AI Research}

\maketitle
In the supplementary material, we first provide more details about the curriculum learning of perturbation (with pseudo-code provided). Then, for all experiments, we present the numbers corresponding to the curves shown in the main paper so that one can easily compare any approach to ours. We also perform an ablation study of the effect of $K$ when evaluating our approach on Human3.6M for long-term motion prediction in the standard setting (i.e., 50 observed frames and 25 predicted frames).

\section{Curriculum Learning of Perturbation}
As discussed in the main paper, our approach benefits from curriculum perturbation of hidden states. The parameter $\alpha$ determines the trade-off between the level of determinism and the quality of motions. In Algorithm 1, we provide the pseudo-code for our Mix-and-Match index sampling.

\begin{algorithm}[H]
\scriptsize
\SetAlgoLined
\textbf{Inputs:} current epoch \texttt{e}, perturbation ratio $\alpha$, sampling step \texttt{step}

\KwResult{Sampled indices for perturbation at epoch \texttt{e}, \texttt{indices}}

$s$ = int($\frac{\texttt{e}}{\texttt{step}}$)

$th$ = $min\{\frac{\alpha L}{2}, \frac{L-\alpha L}{2}\}$

\texttt{indices} = []

\If{$s<th$}
{
\texttt{indices}.append(sample $\alpha L - s$ indices from known part)

\texttt{indices}.append(sample $s$ indices from the rest)
}
\Else
{
\texttt{indices}.append(sample $\alpha L$ indices from hidden state)
}

\Return \texttt{indices}

 \caption{Curriculum Index Sampling}
 \label{alg:curriculum}
\end{algorithm}

\section{Human Evaluation of Quality}
As discussed in the main paper, when evaluating quality with a classifier, the classifier may \emph{memorize} the training set and thus become biased toward real data. As such, this would penalize the motions generated by our model, which are diverse enough to contain motions that were not seen during training. Therefore, human evaluation nicely complements the classifier-based quality assessment. In Table~\ref{tab:huamn}, we provide statistics related to the classifier-based and human evaluation of the quality of motions.

\begin{table}[!t]
    \centering
    \begin{tabular}{l c c}
         Baseline & $\texttt{Q}_\texttt{Classifier}$ & $\texttt{Q}_\texttt{Human}$\\
         \hline
         \texttt{LHP}~\cite{yan2018mt} & 45.0 & 37.0\\
         \texttt{RHP}~\cite{barsoum2018hp} & 47.0 & 38.1\\
         \texttt{LPP}~\cite{walker2017pose} & 13.0 & 15.0\\
         \hline
         \texttt{Ours} & 42.0 & 40.9\\
    \end{tabular}
    \caption{Comparing $\texttt{Q}_\texttt{Classifier}$ and human evaluation of quality for our approach and the baselines. These statistics correspond to evaluation after the models are fully trained.}
    \label{tab:huamn}
\end{table}

\section{Quality and Diversity of Motions}
In Table~\ref{tab:QD}, we provide the quality and corresponding diversity measures for the motions generated by our approach as well as the stochastic baselines compared with.

\begin{table*}[h]
\scriptsize
    \centering
    \begin{tabular}{l c c c c c c c c c c}
    \hline
     & \multicolumn{10}{c}{Training Progress}\\
    \hline
    Baseline & 10\% & 20\% & 30\% & 40\% & 50\% & 60\% & 70\% & 80\% & 90\% & 100\%\\
    \hline
        \texttt{LHP}~\cite{yan2018mt} & 1.19/29.0 & 1.13/37.0 & 0.28/45.0 & 0.24/46.0 & 0.21/46.0 & 0.47/45.0 & 0.28/44.0 & 0.26/45.0 & 0.24/46.0 & 0.26/45.0\\
        \texttt{RHP}~\cite{barsoum2018hp} & 0.76/27.0 & 0.65/40.0 & 0.62/43.0 & 0.58/45.0 & 0.54/44.0 & 0.52/44.0 & 0.51/45.0 & 0.50/45.0 & 0.48/46.0 & 0.48/47.0 \\
        \texttt{LPP}~\cite{walker2017pose}& 2.21/03.0 & 1.35/11.0 & 1.23/12.0 & 1.28/12.0 & 1.43/14.0 & 1.58/11.0 & 1.69/12.0 & 1.74/14.0 & 1.75/14.0 & 1.70/13.0 \\
        \hline
        \texttt{Ours} &
         2.02/22.0 &
         2.34/27.0 &
         2.73/30.0 &
         2.95/34.0 &
         3.17/39.0 &
         3.26/40.0 &
         3.49/41.0 &
         3.63/42.0 &
         3.53/42.0 &
         3.52/42.0 \\
        \hline
    \end{tabular}
    \caption{Quality and diversity of our approach and the stochastic baselines. The results are provided in the form of $Diversity/Quality$ at different training epochs.}
    \label{tab:QD}
\end{table*}

\section{Evaluating the Effect of $K$}
In the main paper, we used $K=50$ to compare our approach with the state-of-the-art deterministic and stochastic baselines. Here, we provide an ablation study on the effect of $K$. To this end, we provide results when using $K=1$ to $K=500$. In Fig.~\ref{fig:k_effect}, we plot the results with $K=50$ as bold black lines, and the shaded area covers the results obtained with $K=1$ to $K=500$. While smaller values of $K$ yield large errors, the difference between $K=50$ and $K=500$ is very small (barely visible in most cases). 
\begin{figure*}[!t]
    \centering
    \begin{tabular}{c@{ }c@{ }c@{ }c}
    
    \pgfplotsset{width=0.27\textwidth,height=0.18\textheight, compat=1.15, grid style={dashed,gray}}  

\begin{tikzpicture}
\tiny
\begin{axis}[
  ymin=0.3, ymax=1.,
   xlabel=Time (ms),
   title={Walking},
ylabel={\tiny Error (MAE)},
  grid=both,
  minor tick num=1
]
\scriptsize

\addplot[smooth,black, style={thick}]
  coordinates{
(	80	,	0.33	)
(	160	,	0.48	)
(	320	,	0.56	)
(	400	,	0.58	)
(	560	,	0.64	)
(	1000	,	0.68	)

}; \label{Ours}

\addplot[name path=plot1, smooth,gray]
  coordinates{
(	80	,	0.34	)
(	160	,	0.50	)
(	320	,	0.58	)
(	400	,	0.63	)
(	560	,	0.68	)
(	1000	,	0.74	)

}; \label{Ours min}

\addplot[name path=plot2, smooth,gray]
  coordinates{
(	80	,	0.33	)
(	160	,	0.47	)
(	320	,	0.56	)
(	400	,	0.57	)
(	560	,	0.63	)
(	1000	,	0.67	)

}; \label{Ours max}

\addplot[fill=gray, opacity=0.4] fill between[of = plot1 and plot2];
\end{axis}

\end{tikzpicture}  & 
    \pgfplotsset{width=0.27\textwidth,height=0.18\textheight, compat=1.3, legend columns=1,  legend style={at={(-0.52,1)},anchor=north}, grid style={dashed,gray}}  

\begin{tikzpicture}
\tiny
\begin{axis}[
  ymin=0.2, ymax=1.1,
   xlabel=Time (ms),
   title={Eating},
  grid=both,
  minor tick num=1
]
\scriptsize

\addplot[smooth,black, style={thick}]
  coordinates{
(	80	,	0.23	)
(	160	,	0.34	)
(	320	,	0.41	)
(	400	,	0.5	)
(	560	,	0.61	)
(	1000	,	0.91	)

}; \label{Ours}

\addplot[name path=plot1, smooth,gray]
  coordinates{
(	80	,	0.23	)
(	160	,	0.36	)
(	320	,	0.44	)
(	400	,	0.53	)
(	560	,	0.64	)
(	1000	,	0.96	)

}; \label{Ours min}

\addplot[name path=plot2, smooth,gray]
  coordinates{
(	80	,	0.23	)
(	160	,	0.34	)
(	320	,	0.40	)
(	400	,	0.49	)
(	560	,	0.60	)
(	1000	,	0.90	)

}; \label{Ours max}

\addplot[fill=gray, opacity=0.4] fill between[of = plot1 and plot2];
\end{axis}

\end{tikzpicture}  & 
    \pgfplotsset{width=0.27\textwidth,height=0.18\textheight, compat=1.3, legend columns=1,  legend style={at={(-0.52,1)},anchor=north}, grid style={dashed,gray}}

\begin{tikzpicture}
\tiny
\begin{axis}[
  ymin=0.2, ymax=1.5,
   xlabel=Time (ms),
   title={Smoking},
  grid=both,
  minor tick num=1
]
\scriptsize

\addplot[smooth,black, style={thick}]
  coordinates{
(	80	,	0.23	)
(	160	,	0.42	)
(	320	,	0.79	)
(	400	,	0.77	)
(	560	,	0.82	)
(	1000	,	1.25	)

}; \label{Ours}

\addplot[name path=plot1, smooth,gray]
  coordinates{
(	80	,	0.23	)
(	160	,	0.44	)
(	320	,	0.81	)
(	400	,	0.83	)
(	560	,	0.89	)
(	1000	,	1.33	)

}; \label{Ours min}

\addplot[name path=plot2, smooth,gray]
  coordinates{
(	80	,	0.22	)
(	160	,	0.42	)
(	320	,	0.79	)
(	400	,	0.77	)
(	560	,	0.81	)
(	1000	,	1.24	)

}; \label{Ours max}

\addplot[fill=gray, opacity=0.4] fill between[of = plot1 and plot2];
\end{axis}

\end{tikzpicture}  & 
    \pgfplotsset{width=0.27\textwidth,height=0.18\textheight, compat=1.3, legend columns=1,  legend style={at={(-0.52,1)},anchor=north}, grid style={dashed,gray}}  
\begin{tikzpicture}
\tiny
\begin{axis}[
  ymin=0.2, ymax=1.5,
   xlabel=Time (ms),
   title={Discussion},
  grid=both,
  minor tick num=1
]
\scriptsize

\addplot[smooth,black, style={thick}]
  coordinates{
(	80,		0.25	)
(	160	,	0.60	)
(	320	,	0.83	)
(	400	,	0.89	)
(	560	,	1.12	)
(	1000	,	1.30	)

}; \label{Ours}

\addplot[name path=plot1, smooth,gray]
  coordinates{
(	80,		0.25	)
(	160	,	0.62	)
(	320	,	0.85	)
(	400	,	0.92	)
(	560	,	1.16	)
(	1000	,	1.36	)

}; \label{Ours min}

\addplot[name path=plot2, smooth,gray]
  coordinates{
(	80,		0.25	)
(	160	,	0.60	)
(	320	,	0.83	)
(	400	,	0.88	)
(	560	,	1.09	)
(	1000	,	1.29	)

}; \label{Ours max}

\addplot[fill=gray, opacity=0.4] fill between[of = plot1 and plot2];
\end{axis}

\end{tikzpicture} \\
    \end{tabular}
    \caption{Effect of $K$ in the MAE of our actions of the Human3.6M dataset. Note, the bold black one is the best of $K=50$ motions, and the shaded area indicates the region between best of $K=1$ and $K=500$.}
    \label{fig:k_effect}
\end{figure*}

\section{Long-term Motion Prediction}

In the main paper, we provide the plots of human motion predictions up to 1000ms for our approach and the recent deterministic and stochastic baselines on four actions of the Human3.6M dataset.  Note that, as discussed in the main paper, for our approach and the stochastic baselines, we report the best of $K=50$ motions. The numbers associated with this experiment are provided in Table~\ref{tab:det}.

\begin{table*}[!t]
    \centering
    \scriptsize
    \begin{tabular}{c c}
        \begin{tabular}{l| c c c c c c}
    \hline
    \multicolumn{7}{c}{Walking}\\
    \hline
    Method & 80ms  & 160ms & 320ms & 400ms & 560ms & 1000ms \\
    \hline
    Zero Velocity & 
    0.39 & 0.86 & 0.99 & 1.15 & 1.35 & 1.32 \\
    AGED~\cite{gui2018adversarial} & 
    0.22 & 0.36 & 0.55 & 0.67 & 0.78 & 0.91 \\
    LSTM-3LR~\cite{fragkiadaki2015recurrent} & 
    1.18 & 1.50 & 1.67 & 1.76 & 1.81 & 2.20 \\
    SRNN~\cite{jain2016structural} & 
    1.08 & 1.34 & 1.60 & 1.80 & 1.90 & 2.13 \\
    DAE-LSTM~\cite{ghosh2017learning} & 
    1.00 & 1.11 & 1.39 & 1.48 & 1.55 & 1.39 \\
    GRU~\cite{martinez2017human} & 
    0.28 & 0.49 & 0.72 & 0.81 & 0.93 & 1.03 \\
        \hline
        \hline
    \texttt{LHP}~\cite{yan2018mt} & 
    0.73 & 0.79 & 0.90 & 0.93 & 0.95 & 1.05 \\
    \texttt{RHP}~\cite{barsoum2018hp} & 
    0.61 & 0.62 & 0.71 & 0.79 & 0.83 & 1.07 \\
    \texttt{LPP}~\cite{walker2017pose} & 
    0.56 & 0.66 & 0.98 & 1.05 & 1.28 & 1.60 \\
    \hline
    Ours & 
    0.33 & 0.48 & 0.56 & 0.58 & 0.64 & 0.68 \\
    \hline
    \end{tabular} &  
    \begin{tabular}{l| c c c c c c}
    \hline
    \multicolumn{7}{c}{Eating}\\
    \hline
    Method & 80ms  & 160ms & 320ms & 400ms & 560ms & 1000ms \\
    \hline
    Zero Velocity & 
    0.27 & 0.48 & 0.73 & 0.86 & 1.04 & 1.38 \\
    AGED~\cite{gui2018adversarial} & 
    0.17 & 1.28 & 0.51 & 0.64 & 0.86 & 0.93 \\
    LSTM-3LR~\cite{fragkiadaki2015recurrent} & 
    1.36 & 1.79 & 2.29 & 2.42 & 2.49 & 2.82 \\
    SRNN~\cite{jain2016structural} & 
    1.35 & 1.71 & 2.12 & 2.21 & 2.28 & 2.58 \\
    DAE-LSTM~\cite{ghosh2017learning} & 
    1.31 & 1.49 & 1.86 & 1.89 & 1.76 & 2.01 \\
    GRU~\cite{martinez2017human} & 
    0.23 & 0.39 & 0.62 & 0.76 & 0.95 & 1.08 \\
    \hline
    \hline
    \texttt{LHP}~\cite{yan2018mt} & 
    0.68 & 0.74 & 0.95 & 1.00 & 1.03 & 1.38 \\
    \texttt{RHP}~\cite{barsoum2018hp} & 
    0.53 & 0.67 & 0.79 & 0.88 & 0.97 & 1.12 \\
    \texttt{LPP}~\cite{walker2017pose} & 
    0.44 & 0.60 & 0.71 & 0.84 & 1.05 & 1.54 \\
    \hline
    Ours & 
    0.23 & 0.34 & 0.41 & 0.50 & 0.61 & 0.91 \\
    \hline
    \end{tabular}
    \\
    \begin{tabular}{l| c c c c c c}
    \hline
    \multicolumn{7}{c}{Smoking}\\
    \hline
    Method & 80ms  & 160ms & 320ms & 400ms & 560ms & 1000ms \\
    \hline
    Zero Velocity & 
    0.26 & 0.48 & 0.97 & 0.95 & 1.02 & 1.69 \\
    AGED~\cite{gui2018adversarial} & 
    0.27 & 0.43 & 0.82 & 0.84 & 1.06 & 1.21 \\
    LSTM-3LR~\cite{fragkiadaki2015recurrent} & 
    2.05 & 2.34 & 3.10 & 3.18 & 3.24 & 3.42 \\
    SRNN~\cite{jain2016structural} & 
    1.90 & 2.30 & 2.90 & 3.10 & 3.21 & 3.23 \\
    DAE-LSTM~\cite{ghosh2017learning} & 
    0.92 & 1.03 & 1.15 & 1.25 & 1.38 & 1.77 \\
    GRU~\cite{martinez2017human} & 
    0.33 & 0.61 & 1.05 & 1.15 & 1.25 & 1.50 \\
    \hline
    \hline
    \texttt{LHP}~\cite{yan2018mt} & 
    1.00 & 1.14 & 1.43 & 1.44 & 1.68 & 1.99 \\
    \texttt{RHP}~\cite{barsoum2018hp} & 
    0.64 & 0.78 & 1.05 & 1.12 & 1.64 & 1.84 \\
    \texttt{LPP}~\cite{walker2017pose} & 
    0.59 & 0.83 & 1.25 & 1.36 & 1.67 & 2.03 \\
    \hline
    Ours & 
    0.23 & 0.42 & 0.79 & 0.77 & 0.82 & 1.25 \\
\hline
    \end{tabular}     & 
    \begin{tabular}{l| c c c c c c}
    \hline
    \multicolumn{7}{c}{Discussion}\\
    \hline
   Method & 80ms  & 160ms & 320ms & 400ms & 560ms & 1000ms \\
    \hline
    Zero Velocity & 
    0.31 & 0.67 & 0.94 & 1.04 & 1.41 & 1.96 \\
    AGED~\cite{gui2018adversarial} & 
    0.27 & 0.56 & 0.76 & 0.83 & 1.25 & 1.30 \\
    LSTM-3LR~\cite{fragkiadaki2015recurrent} & 
    2.25 & 2.33 & 2.45 & 2.46 & 2.48 & 2.93 \\
    SRNN~\cite{jain2016structural} & 
    1.67 & 2.03 & 2.20 & 2.31 & 2.39 & 2.43 \\
    DAE-LSTM~\cite{ghosh2017learning} & 
    1.11 & 1.20 & 1.38 & 1.42 & 1.53 & 1.73 \\
    GRU~\cite{martinez2017human} & 
    0.31 & 0.68 & 1.01 & 1.09 & 1.43 & 1.69 \\
    \hline
    \hline
    \texttt{LHP}~\cite{yan2018mt} & 
    0.80 & 1.01 & 1.22 & 1.35 & 1.56 & 1.69 \\
    \texttt{RHP}~\cite{barsoum2018hp} & 
    0.79 & 1.00 & 1.12 & 1.29 & 1.43 & 1.71 \\
    \texttt{LPP}~\cite{walker2017pose} & 
    0.73 & 1.10 & 1.33 & 1.34 & 1.45 & 1.85 \\
    \hline
    Ours & 
    0.25 & 0.60 & 0.83 & 0.89 & 1.12 & 1.30 \\
\hline
    \end{tabular}
    
    \end{tabular}
    \caption{Comparison against the state of the art for four actions of the Human3.6M dataset.}
    \label{tab:det}
\end{table*}

{\small
\bibliographystyle{ieee}
\bibliography{egbib}
}